\documentclass[conference]{IEEEtran}
\IEEEoverridecommandlockouts
\usepackage{hyperref}
\usepackage{cite}
\usepackage{multirow}
\usepackage{amsmath,amssymb,amsfonts}
\usepackage{algorithmic}
\usepackage{graphicx}
\usepackage{textcomp}
\usepackage{mathtools}
\usepackage{multicol}
\usepackage{xcolor}
\usepackage{multicol}
\usepackage{comment}

\usepackage{booktabs}
\usepackage[accsupp]{axessibility}  
\usepackage{soul}
\usepackage{bm}
\usepackage{subcaption}

\usepackage{pgf}
\usepackage{tikz}
\usepackage{tikzscale}
\usetikzlibrary{arrows,external}
\usetikzlibrary{spy}
\usepackage{pgfplots}
\usepackage[skins]{tcolorbox}
\usepackage{import}
\usepackage{url}
\usepackage{tabu}

\def\BibTeX{{\rm B\kern-.05em{\sc i\kern-.025em b}\kern-.08em
    T\kern-.1667em\lower.7ex\hbox{E}\kern-.125emX}}
\begin{document}
\title{Enhancing Sketch Animation: Text-to-Video Diffusion Models with Temporal Consistency and Rigidity Constraints}
\author{Gaurav Rai, Ojaswa Sharma
\IEEEauthorblockA{ \\
Graphics Research Group, Indraprastha Institute of Information Technology Delhi\\
gauravr@iiitd.ac.in, ojaswa@iiitd.ac.in}
}
\maketitle
\begin{abstract}
Animating hand-drawn sketches using traditional tools is challenging and complex. Sketches provide a visual basis for explanations, and animating these sketches offers an experience of real-time scenarios. We propose an approach for animating a given input sketch based on a descriptive text prompt. Our method utilizes a parametric representation of the sketch's strokes. Unlike previous methods, which struggle to estimate smooth and accurate motion and often fail to preserve the sketch's topology, we leverage a pre-trained text-to-video diffusion model with SDS loss to guide the motion of the sketch's strokes.
We introduce length-area (LA) regularization to ensure temporal consistency by accurately estimating the smooth displacement of control points across the frame sequence. Additionally, to preserve shape and avoid topology changes, we apply a shape-preserving As-Rigid-As-Possible (ARAP) loss to maintain sketch rigidity. Our method surpasses state-of-the-art performance in both quantitative and qualitative evaluations.
\end{abstract}
\section{Introduction}

\label{sec:introduction}

\begin{figure*}[htbp]
  \centering
  \includegraphics[width=\textwidth]{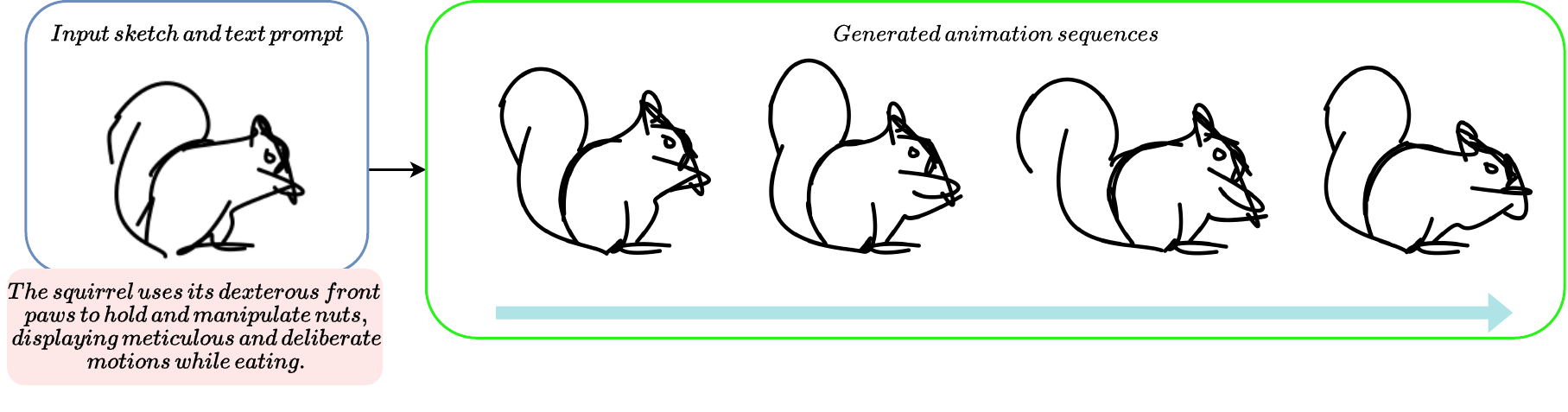}
  \caption{Problem with the LiveSketch~\cite{gal2023breathing} method. In this example, we can observe the lack of temporal consistency and shape distortion during motion.}
  \label{fig:problem}
\end{figure*}

Sketches serve as a medium for communication and visual representation. Animating 2D sketch illustrations using traditional tools is tedious, cumbersome, and requires significant time and effort. Keyframe-based animation is highly labor-intensive, while video-driven animation methods are often restricted to specific motions. In recent years, sketch animation has emerged as a significant area of research in computer animation, with applications in video editing, entertainment, e-learning, and visual representation.
Previous sketch animation methods, such as that in~\cite{xing2015autocomplete,patel2016tracemove}, require extensive manual input and artistic skill, presenting challenges for novice users. Traditional methods are limited to specific types of motion, such as facial and biped animation. More recent techniques, like Su et al.~\cite{su2018live}, animate sketches based on a video, but still require manual input.
AnimationDrawing~\cite{smith2023method} is a sketch animation technique that does not rely on manual input, generating animations using pose mapping, but is limited to biped motion.
In contrast, LiveSketch~\cite{gal2023breathing} is a learning-based approach that takes a sketch and text prompt to produce an animated sketch. While it generates promising results, it faces challenges with temporal consistency and shape preservation (see Figure~\ref{fig:problem}). 
To address these issues, we propose a method for animating input sketches based solely on a text description, with no manual input required. Our approach represents each stroke as a B\'ezier curve, similar to LiveSketch~\cite{gal2023breathing}, and extends LiveSketch's capabilities by a novel Length-Area regularization and rigidity loss. Furthermore, we utilize local and global paths for motion estimation and apply Score Distillation Sampling (SDS) loss~\cite{poole2022dreamfusion} for optimization. 
We propose length-area (LA) regularization that maintains temporal consistency, yielding smooth and accurate motion in the animated sketch. Further, the As-Rigid-As-Possible (ARAP) loss~\cite{igarashi2005rigid} preserves local rigidity in the sketch's shape during animation. Our method outperforms state-of-the-art techniques in both quantitative and qualitative evaluations. We achieve better sketch-to-video consistency and text-to-video alignment compared to previous method. Our main contributions are as follows: 
\begin{itemize} 
    \item We propose a Length-Area regularization to maintain temporal consistency across animated sequences. It allows for the generation of a smooth animation sequence. 
    \item A shape-preserving ARAP loss to preserve local rigidity in sketch strokes during animation. The rigidity loss overcomes the shape distortion during animation. 
\end{itemize}

\section{Related work}
\subsection{Sketch Animation}
Traditional sketch animation tools are time-consuming and require a certain level of artistic skills. Agarwala et al.~\cite{agarwala2004keyframe} proposed a rotoscoping approach that estimates motion from contour tracking and animates the sketches. It reduces manual user inputs in the contour-tracking process. Bregler et al.~\cite{bregler2002turning} extract motion from cartoon animated characters and retarget these to sketches using the keyframe-based approach, producing more expressive results but requires additional user inputs. Guay et al.~\cite{guay2015space} propose a method that enables timestep shape deformation by sketching a single stroke but is limited to a few animation styles. Autocomplete methods~\cite{wang2004video,xing2015autocomplete} predict the subsequent sketching style by the user using temporal coherency, but these methods require manual user input for sketching operations for each keyframe. 
Several learning-based and energy optimization-based animation methods have been proposed in recent years. These methods aim to perform animation using video motion, text prompt input, and predefined motion trajectory. Santosa et al.~\cite{santosa2013direct} animate a sketch by marking over the video using optical flow. However, this method suffers in the case of structural differences between the sketch and the video object. 

Deep learning-based methods~\cite{liu2019neuroskinning,jeruzalski2020nilbs,xu2020rignet} provide an alternative for animators by demonstrating robust capacity for rig generation. Animation Drawing~\cite{smith2023method} generates a rigged character of children's drawing using the alpha-pose mapping from a predefined character motion. SketchAnim~\cite{rai2024sketchanim} maps the video skeleton to the sketch skeleton and estimates the skeleton transformation to animate the sketch using skinning weights. It handles self-occlusion and can animate non-living objects but fails to animate stroke-level sketches.
CharacterGAN~\cite{hinz2022charactergan} generates an animation sequence (containing a single character) by training a generative network with only 8-15 training samples with keypoint annotation defined by the user. Neural puppet~\cite{poursaeed2020neural} adapts the animation of hand-drawn characters by providing a few drawings of the characters in defined poses. Video-to-image animation~\cite{siarohin2019first,siarohin2021motion,wang2022latent,mallya2022implicit,tao2022structure,zhao2022thin} methods extract the motion of keypoint learning-based optical flow estimated from the driving video and generate the animated images. However, it is limited to the image modality.
AnaMoDiff~\cite{tanveer2024anamodiff} estimates the optical flow field from a reference video and warps it to the source input. Su et al.~\cite{su2018live} defines control points on the first frame of the video, tracks the control points in the video for all frames, and applies this motion to the control points on the input sketch. 
Unlike previous methods that require skeletons, control points, or reference videos, our approach generates high-quality, non-rigid, smooth sketch deformations using only text prompts without manual user input.

\subsection{Image and text-to-video generation}
Text-to-video generation aims to produce the corresponding video using text prompt input automatically. Previous works have discovered the ability of GANs~\cite{tian2021good,zhu2023motionvideogan,li2018video} and auto-regressive transformers~\cite{wu2021godiva,yan2021videogpt} for video generation, but these are restricted to the fixed domain.
Recent progress in diffusion models sets an enrichment in the video generation methodology. Recent methods such as ~\cite{wang2023modelscope,chen2023videocrafter1,guo2023animatediff,chen2023livephoto,zhou2022magicvideo} utilize Stable Diffusion~\cite{ni2023conditional} to incorporate temporal information in latent space. Dynamicrafter~\cite{xing2023dynamicrafter} generates videos from input images and text prompts.
Despite advancements, open-source video generation faces challenges in maintaining text readability during motion. LiveSketch~\cite{gal2023breathing}  animates vector sketches that do not require extensive training. It uses a pre-trained text-to-video diffusion model to utilize motion and instruct the motion to sketch using SDS~\cite{poole2022dreamfusion}.
Similar to text-to-video generation and image-to-video generation, a closed research area aims to generate video from an input image. Latent Motion Diffusion~\cite{hu2023lamd} estimates the motion by learning the optical flow from video frames and uses the 3D-UNet diffusion model to generate the animated video. Make-It-Move~\cite{hu2022make} uses an encoder-decoder network condition on image and text prompt input to generate the video sequences. VideoCrafter1~\cite{chen2023videocrafter1} and LivePhoto~\cite{chen2023livephoto} preserve the input image style and structure by training to be conditioned on text and image input. CoDi~\cite{tang2024any} trained on shared latent space with conditioning and output space and aligned modalities such as image, video, text, and audio. However, these approaches struggle to preserve the characteristics of the vectorized input sketch.

DreamFusion~\cite{poole2022dreamfusion} proposed SDS loss that generates 3D representations from text input using 2D image diffusion. The SDS loss is similar to diffusion model loss. However, it does not include the U-Net jacobian, which helps it overcome the high computation time of backpropagation within the diffusion model and aligns the image per the text condition by guiding the optimization process. SDS loss is also used to optimize the other generative tasks such as sketches~\cite{xing2024diffsketcher}, vector graphics~\cite{jain2023vectorfusion}, and meshes~\cite{chen2023fantasia3d}.
The diffusion network predicts the sketch points' position for each frame and aligns the entire animation with the text prompt using SDS loss.

\section{Methodology}
\begin{figure*}[!htbp]
  \centering
  \includegraphics[width=0.9\textwidth]{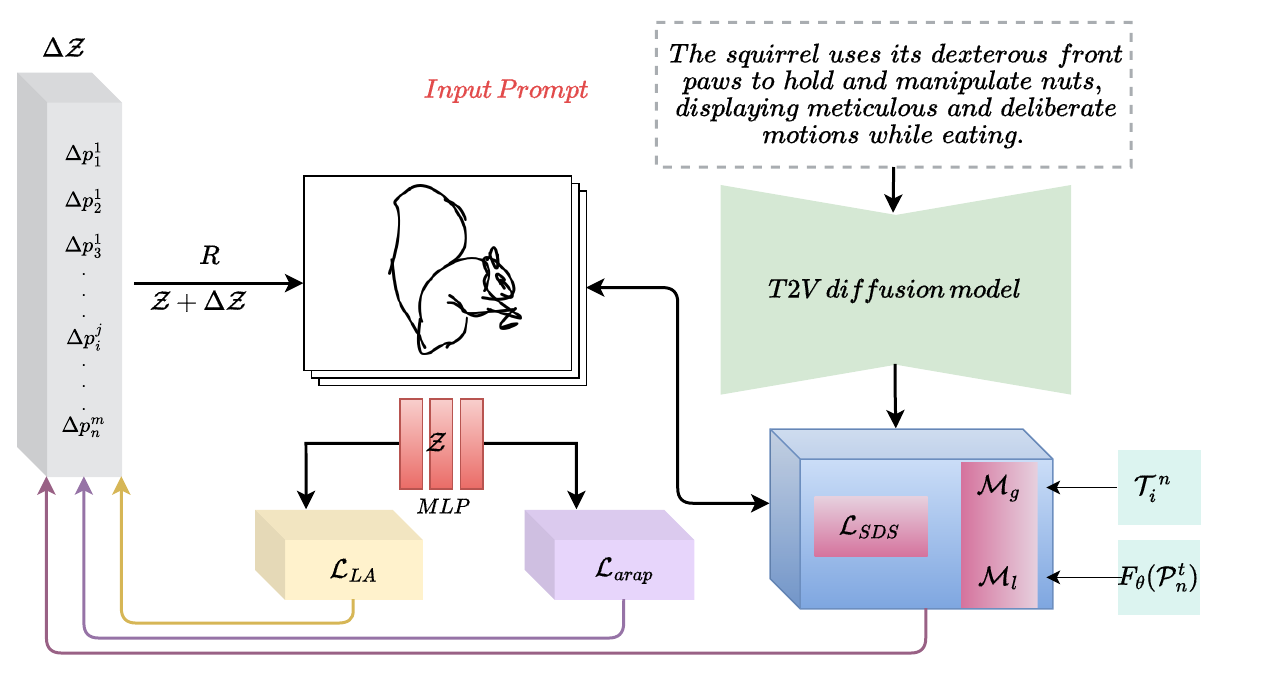}
  \caption{Network architecture of our proposed framework. We use a Length-Area regularizer to maintain temporal consistency and avoid drastic shape changes. ARAP loss maintains the sketch stroke's rigidity and prevents shape distortions during motion.}
  \label{fig:network}
\end{figure*}

Our methodology extends the framework introduced by Gal et al.~\cite{gal2023breathing}, which produces animations from sketches guided by textual descriptions. Each sketch consists of a set of strokes, represented as cubic B\'ezier curves. We represent the set of control points within a frame as $\mathrm{B} = \left\{ { \mathbf{p}_i } \right\}_{i=1}^{4k}$ where $\mathbf{p}_{i}\in\mathbb{R}^2$ and $k$ is the total number of strokes. Further, we define a sketch video of $n$ frames by a set of moving control points $\mathrm{Z} = \left\{ { \mathrm{B}_i } \right\}_{i=1}^{n}$, where $\mathrm{Z}\in \mathbb{R}^{4k\times n\times 2}$.

Animation of a sketch requires the user to provide a text prompt passed into the network as input along with the sketch. Similar to LiveSketch~\cite{gal2023breathing}, we use a neural network architecture that takes an initial set of control points, $\mathrm{Z}^{init}$, as input and produces the corresponding set of displacements, $\Delta \mathrm{Z}$. For each frame, $\mathrm{Z}^{init}$ is initialized to the set $\mathrm{B}$. 
 
Each control point is first projected onto a latent space using a mapping function, $g_{shared} : \mathbb{R}^2 \rightarrow \mathbb{R^D}$. This function takes the initial point set $\mathrm{Z}^{init} \in \mathbb{R}^2$ and projects it into a higher-dimensional space enriched with positional encoding, thereby generating point features. 

These features are processed through two branches: a local motion predictor, $\mathcal{M}_l$, implemented as a multi-layer perceptron (MLP), which computes unconstrained local motion offsets, and a global motion predictor, $\mathcal{M}_g$, which estimates transformation matrices $\mathbf{M}_i$ for scaling, shear, rotation, and translation, yielding the global motion offsets.
The generated animation sequence suffers from a lack of temporal consistency and degradation of sketch identity during motion.
We propose a novel Length-Area (LA) regularization framework to significantly enhance temporal coherence in animated sequences. Our approach estimates the B\'ezier curve length and the area between consecutive frames, optimizing these metrics through a multilayer perceptron (MLP). Furthermore, we introduce a shape-preserving As-Rigid-As-Possible (ARAP) loss, leveraging a mesh constructed via Delaunay triangulation~\cite{delaunay1934} of control points within each frame. Unlike existing methods, our ARAP loss is explicitly designed to maintain local shape consistency, addressing critical challenges in deformation handling and ensuring robust animation fidelity. Figure~\ref{fig:network} provides a detailed illustration of our proposed network architecture, highlighting its key components. To evaluate its performance, we experimented with different learning rate configurations and conducted multiple iterations of the optimization process, systematically refining the model.
 
\subsection{Regularization}
The LA regularizer is designed to minimize abrupt changes in stroke lengths between consecutive frames, ensuring smoother transitions and preserving structural consistency by maintaining stable stroke lengths across the animation. To mitigate error propagation, the length minimization for a stroke in a given frame is computed relative to its length in the initial frame. The stroke length is estimated as the curve length, $L = \int_0^1 |\dot{\mathbf{f}}(u)|\,\mathrm{d}u$ of the B\'ezier curve $\mathbf{f}(u)$.

B\'ezier curves lack local control, meaning that even minor adjustments to control point positions can lead to significant changes in the resulting curve. To mitigate this issue, we introduce an area loss term that minimizes the area spanned by a stroke between consecutive frames, thereby enhancing temporal stability and reducing undesirable deformations. To compute this area, we consider a stroke represented by the B\'ezier curve $\mathbf{f}_i(u)$ defind by the control points $\mathbf{p}_{i,j}$, where $j=\{0\ldots 3\}$, for an intermediate frame $i$. Let the estimated global transformation matrix of control points for frame $i$ be denoted as $\mathbf{M}_i$, and the corresponding local motion offsets as $\Delta\mathbf{p}_{i,j}$. The control points for the next frame are determined as $\mathbf{p}_{i+1,j} = \mathbf{M}_i\mathbf{p}_{i,j} + \Delta\mathbf{p}_{i,j}$. The space-time B\'ezier surface $\mathbf{f}(u, t)$ for $t\in[t_i, t_{i+1}]$ is defined by time-varying control points $\mathbf{p}_{j}(t) = \mathbf{M}(t)\mathbf{p}_{i,j} + \Delta\mathbf{p}_{i,j}(t-t_i)/(t_{i+1}-t_i)$, where $\mathbf{M}(t)$ is obtained by interpolating the transformation parameters appropriately over time. The surface area swept by the stroke between frames $i$ and $i+1$  is computed as
\begin{align}
A_i = \int_{t_i}^{t_{i+1}}\int_0^1 \left\lVert\frac{\partial\mathbf{f}}{\partial u}\times \frac{\partial\mathbf{f}}{\partial t}\right\rVert \mathrm{d}u \,\mathrm{d}t.
\end{align}

The LA regularization, denoted as $\mathcal{L}_{LA}$ is defined as
\begin{equation}
    \label{eqn:la_loss}
    \mathcal{L}_{{LA}} = \sum_{i=0}^{n-1} \left(\lambda_{l} \left| L_{i+1} - L_{i} \right| + \lambda_{a} A_i\right),
\end{equation}
where $\mathcal{L}_{LA}$ represents the length-area loss function. This formulation aims to minimize both the variation in stroke length and the swept area between consecutive frames, ensuring temporal coherence and stability in animation. We use a multilayer perceptron (MLP) to optimize this loss, with values of hyperparameters $\lambda_{l}$ and $\lambda_{a}$ set to $0.1$ and $1e-5$, respectively.

LiveSketch~\cite{gal2023breathing} uses the SDS loss to train its model, which has separate blocks for optimizing the global and local motion. The SDS loss is defined as
\begin{equation}
    \label{eqn:sds}
    \bigtriangledown_{\phi} \mathcal{L}_{sds} = \left [ w(\gamma)(\epsilon_{\theta}(x_{\gamma},\gamma,y) - \epsilon) \frac{\delta x}{\delta \phi}\right ], 
\end{equation}
where $\epsilon_{\theta}(x_{\gamma},\gamma,y)$ is the output of the diffusion model, $\epsilon$ denotes the actual noise, $\gamma$ represents the timestep, and $w(\gamma)$ is a constant term that depends on the noising schedule. The SDS loss is calculated at every step of the diffusion generation process for all frames, guiding the training of these blocks and the overall generation process.
During each generation step, the optimization occurs after completing the SDS loss-based optimization for both blocks of the base model, resulting in updated control points. These control points are used as input for our optimization procedure.

\begin{figure}[!htbp]
  \centering
  \includegraphics[width=0.4\textwidth]{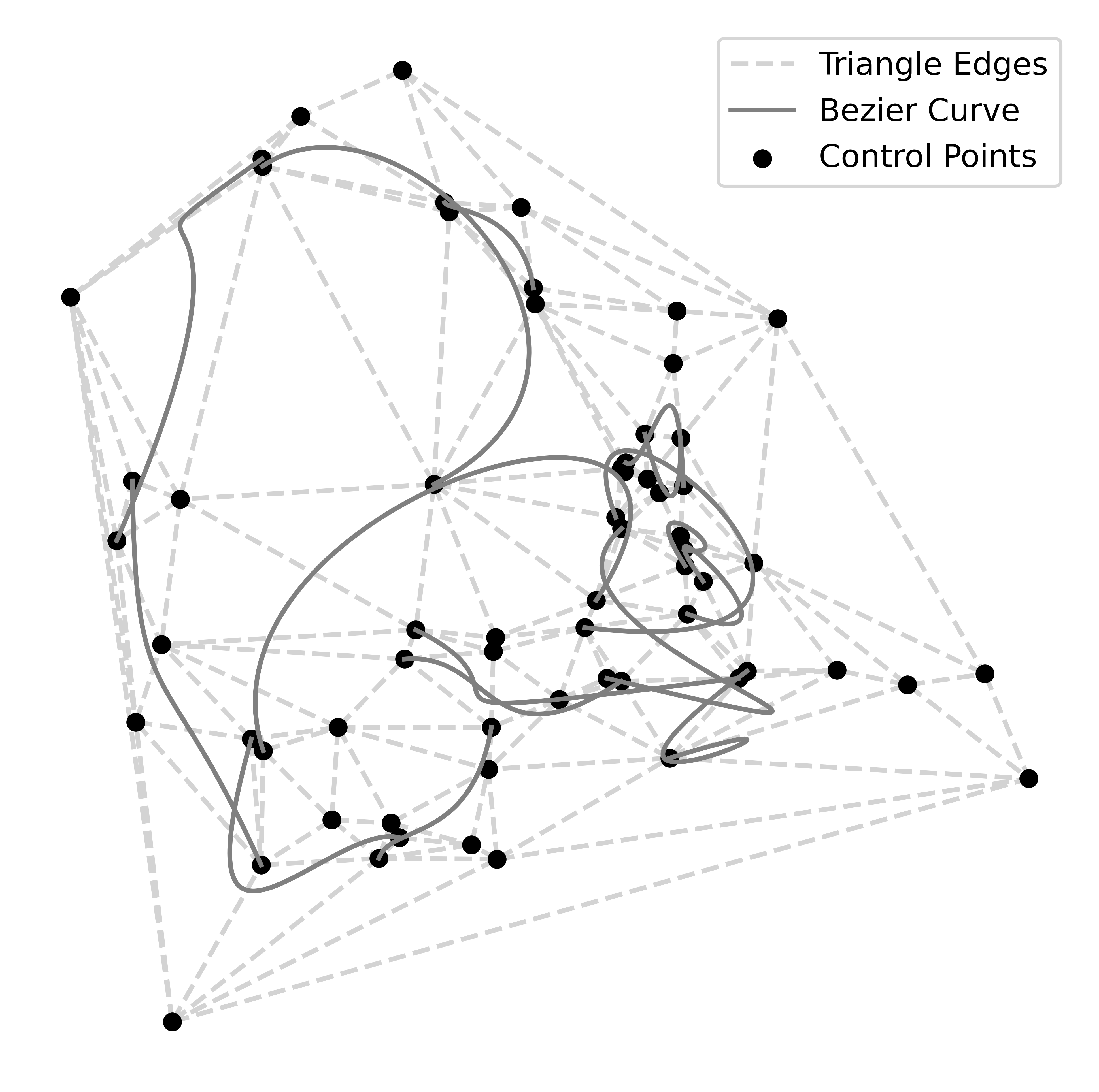}
  \caption{Cubic B\'ezier curve of each stroke and their corresponding control points. Delaunay triangulation of B\'ezier control points.}
  \label{fig:triangle}
\end{figure}

\subsection{Shape preservation}
As-rigid-as-possible deformation enables point-driven shape deformation by moving anchor points, which act as constraints within the model. This deformation framework maintains the rigidity of each element of the mesh as closely as possible, ensuring that transformations are smooth and visually coherent. The ARAP method leverages a two-step optimization algorithm. In the first step, an initial rotation is estimated for each triangle in the mesh. This involves computing the optimal rotation matrix that best approximates the transformation required to map the vertices of each triangle from their initial positions to their target positions while minimizing distortion. 
The second step involves adjusting the scale, ensuring that the transformation adheres to an as-rigid-as-possible model by minimizing the amount of stretch that would distort the original shape. The approach minimizes distortion across the triangular mesh by optimizing each triangle's local transformations while maintaining global consistency across the mesh.

In our proposed approach, we extend the standard ARAP loss~\cite{igarashi2005rigid} by formulating it as a differentiable function, enabling the use of gradient-based optimization techniques and backpropagation within the network. This differentiable ARAP loss is optimized using a multilayer perceptron (MLP), allowing adaptive and flexible shape deformation.

The ARAP loss is computed based on a global mesh structure formed by triangulating B\'ezier control points (see Figure~\ref{fig:triangle}) within each frame. Calculating the ARAP loss relative to a similar triangulated mesh for the next frame ensures stroke preservation, which is essential for generating smooth and consistent animations. The ARAP loss $\mathcal{L}_{ARAP}$ is computed by identifying all triangles in the mesh formed by the control points of a given frame, with the triangulation topology $\mathcal{T}$ remaining fixed across all frames. The same is defined as
\begin{align}
\label{eqn:arap}
    \mathcal{L}_{ARAP} = \sum_{e\in\mathcal{T}}\alpha_e\left \|e^\prime  - \mathbf{D}e\right \|^2,
\end{align}
where $\mathbf{D}$ is the ARAP transformation matrix, ${e}$ represents an edge of a triangle, estimated from the control points of the initial sketch, and ${e^\prime}$ denotes the corresponding deformed edge of the triangle of the subsequent frames. $\alpha_e$ denotes the weight, usually proportional to the edge length. The ARAP loss in equation~\ref{eqn:arap} is calculated by first identifying the triangles that form the mesh of the given frame. These triangles are used to compute the transformation matrix, which is then optimized using a multi-layer perceptron (MLP).

\begin{figure*}[!ht]
  \centering
    \setlength{\tabcolsep}{2pt} 
    \begin{tabular}{c|ccccc}
         \textbf{\small Input Sketch} & \multicolumn{5}{c}{$\overbrace{\rule{5in}{0pt}}^\text{\textbf{\small Generated frames}}$} \\
         
         \multirow{2}{*}{
           \begin{minipage}[c]{.15\linewidth}
             \includegraphics[width=1.2\linewidth]{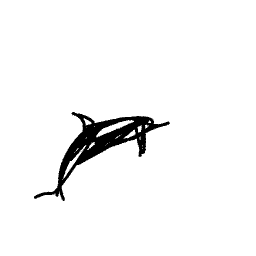}
           \end{minipage}
         } \\ [-2pt]
         & \raisebox{-0.5\height}{\includegraphics[width=.15\linewidth]{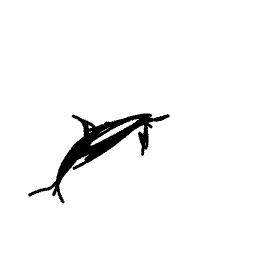}}
         & \raisebox{-0.5\height}{\includegraphics[width=.15\linewidth]{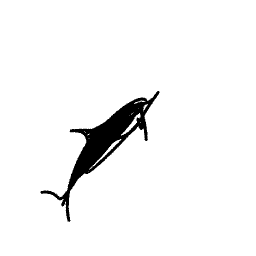}}
         & \raisebox{-0.5\height}{\includegraphics[width=.15\linewidth]{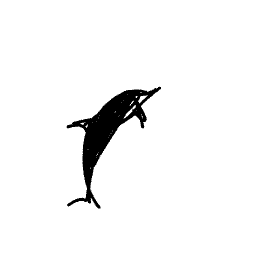}} 
         & \raisebox{-0.5\height}{\includegraphics[width=.15\linewidth]{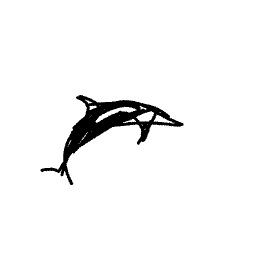}} 
         & \raisebox{-0.5\height}{\includegraphics[width=.15\linewidth]{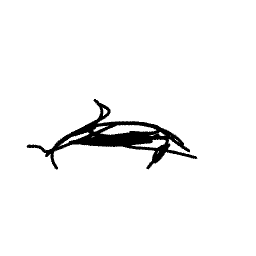}} \\
         & \multicolumn{5}{c}{\small \textcolor{blue}{Text Prompt: } "A dolphin swimming and leaping out of the water"} \\
         
         \multirow{2}{*}{
           \begin{minipage}[c]{.12\linewidth}
             \includegraphics[width=1.2\linewidth]{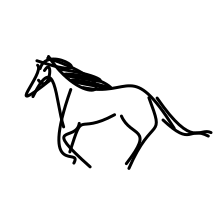}
           \end{minipage}
         }\\ [-12pt]
         & \raisebox{-0.5\height}{\includegraphics[width=.15\linewidth]{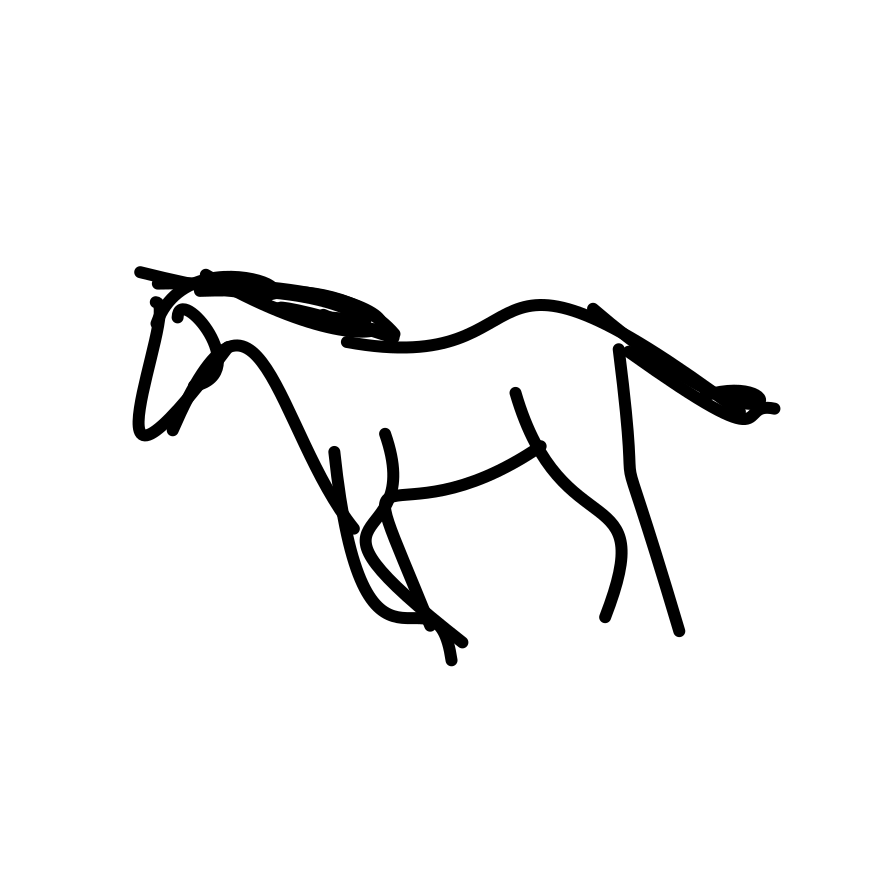}}
         & \raisebox{-0.5\height}{\includegraphics[width=.15\linewidth]{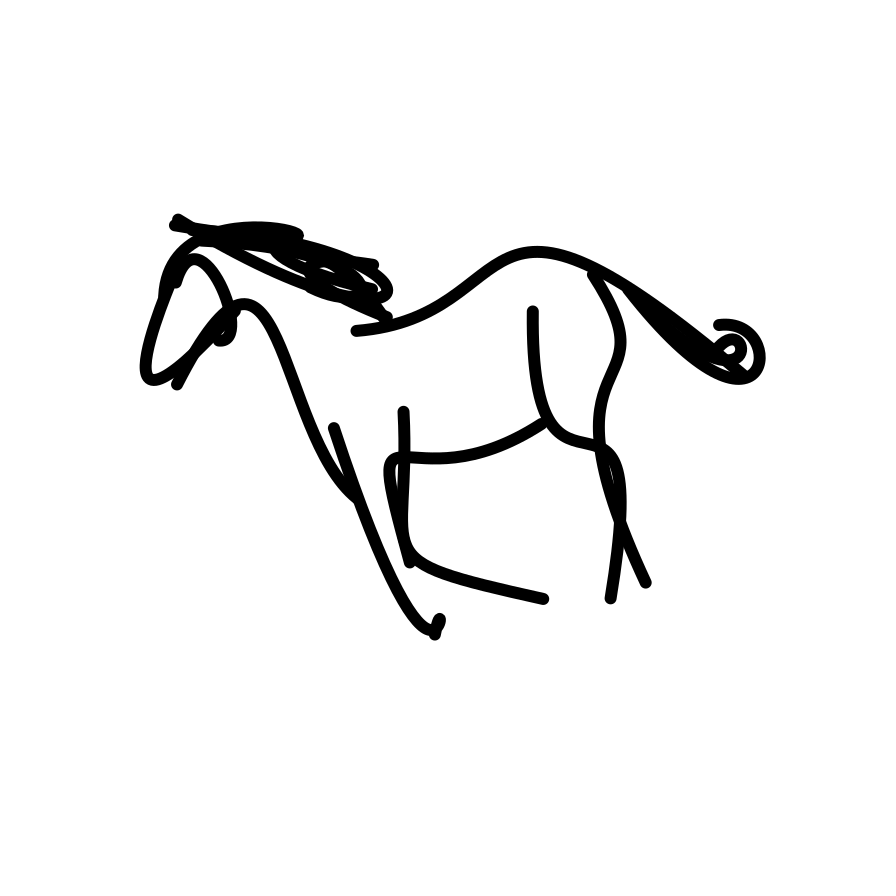}}
         & \raisebox{-0.5\height}{\includegraphics[width=.15\linewidth]{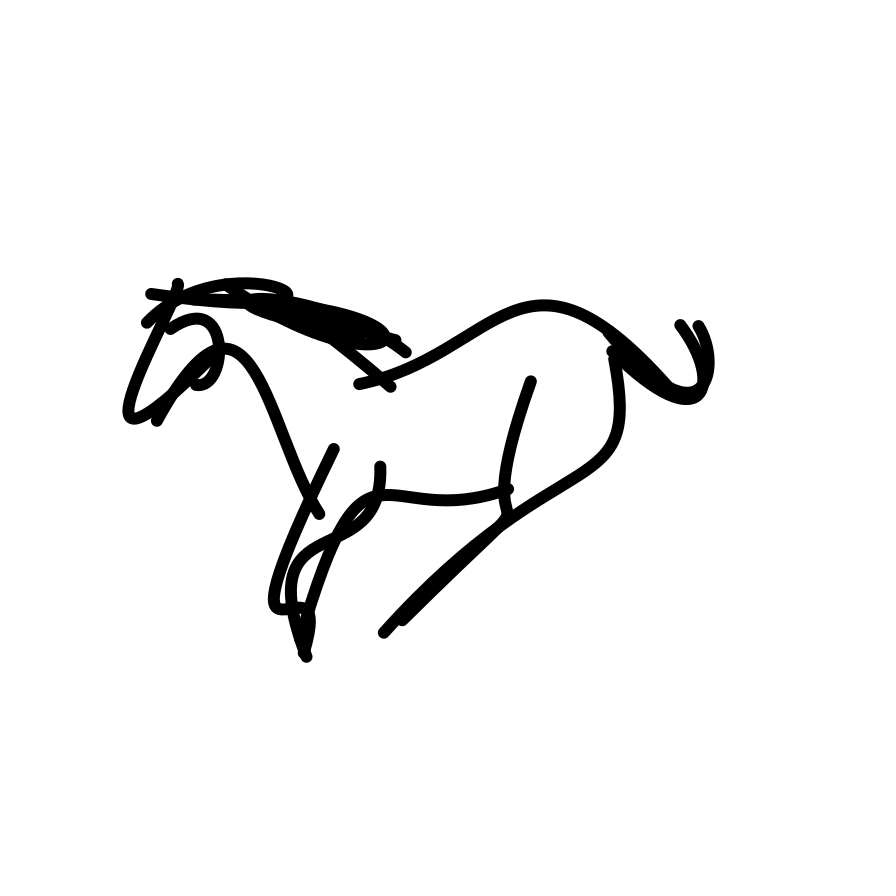}}
         & \raisebox{-0.5\height}{\includegraphics[width=.15\linewidth]{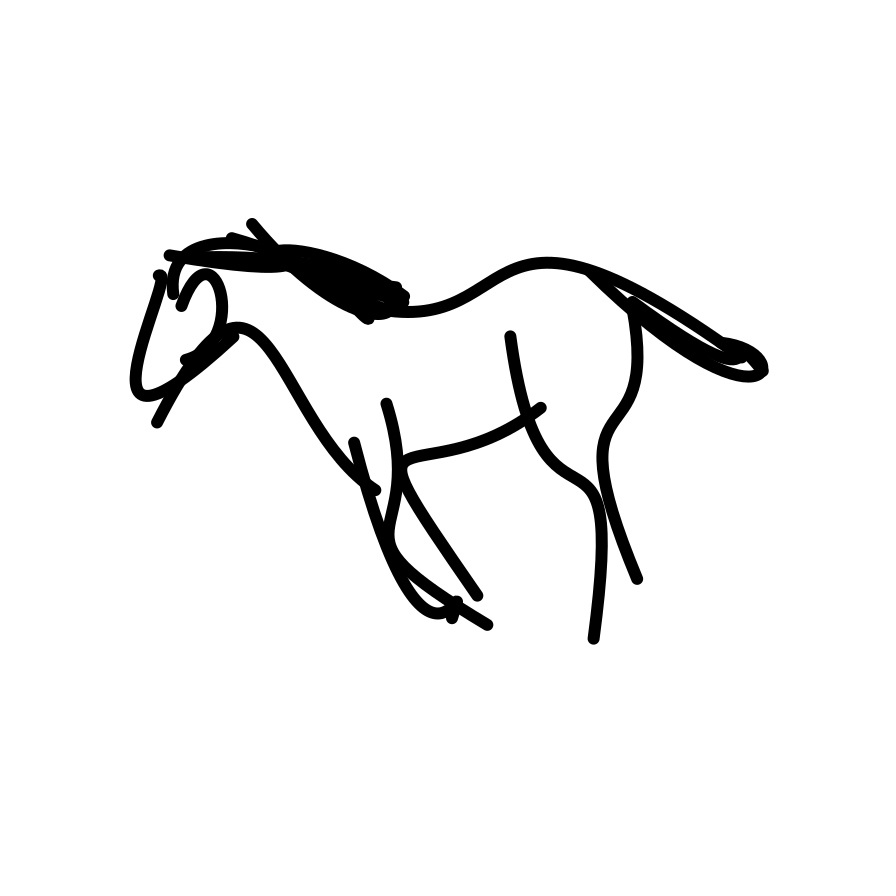}} 
         & \raisebox{-0.5\height}{\includegraphics[width=.15\linewidth]{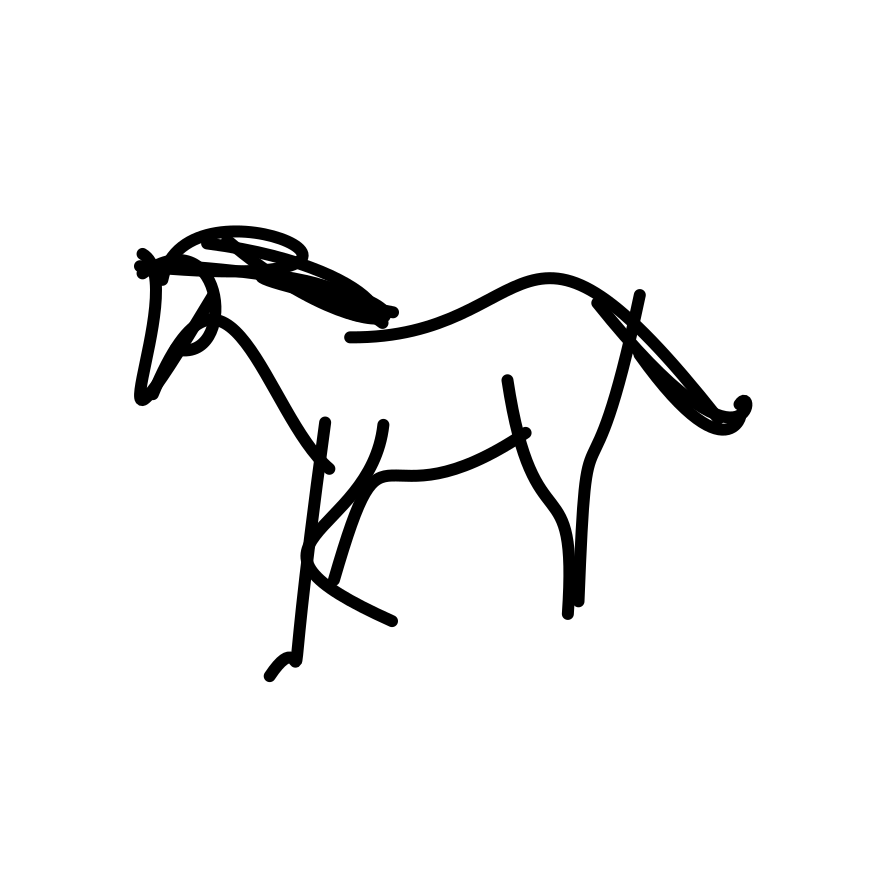}} \\
         & \multicolumn{5}{c}{\small \textcolor{blue}{Text Prompt: } "A galloping horse."} \\
         
         \multirow{2}{*}{
           \begin{minipage}[c]{.12\linewidth}
             \includegraphics[width=\linewidth]{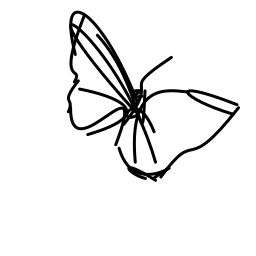}
           \end{minipage}
         }\\ [-12pt]
         & \raisebox{-0.5\height}{\includegraphics[width=.15\linewidth]{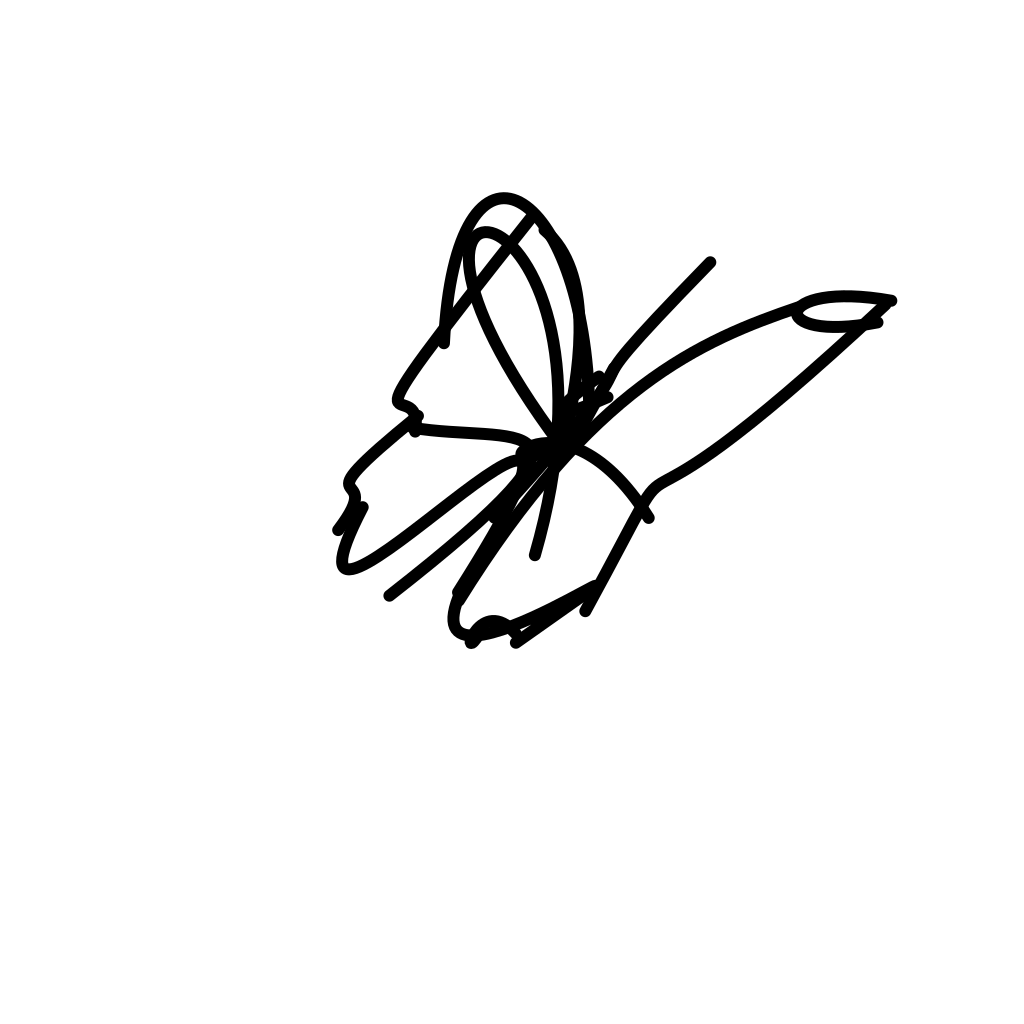}}
         & \raisebox{-0.5\height}{\includegraphics[width=.15\linewidth]{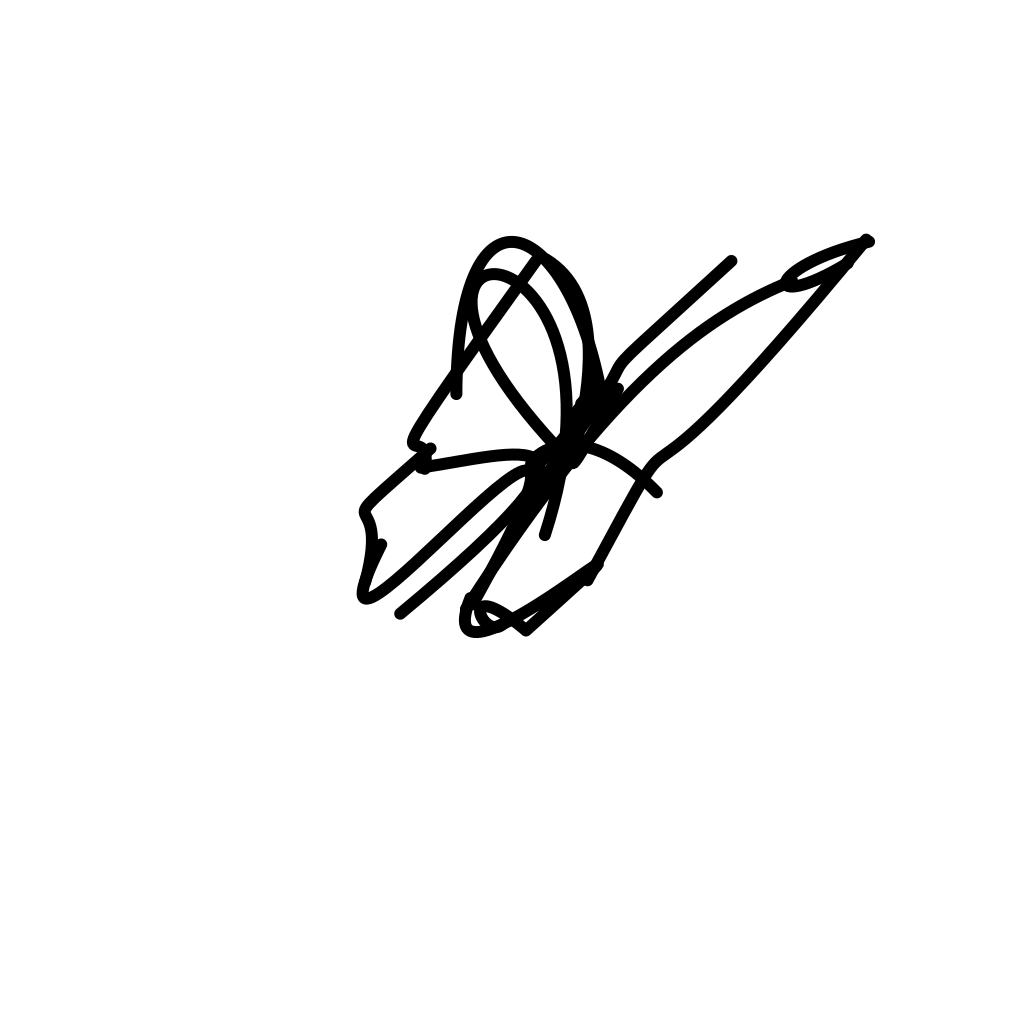}}
         & \raisebox{-0.5\height}{\includegraphics[width=.15\linewidth]{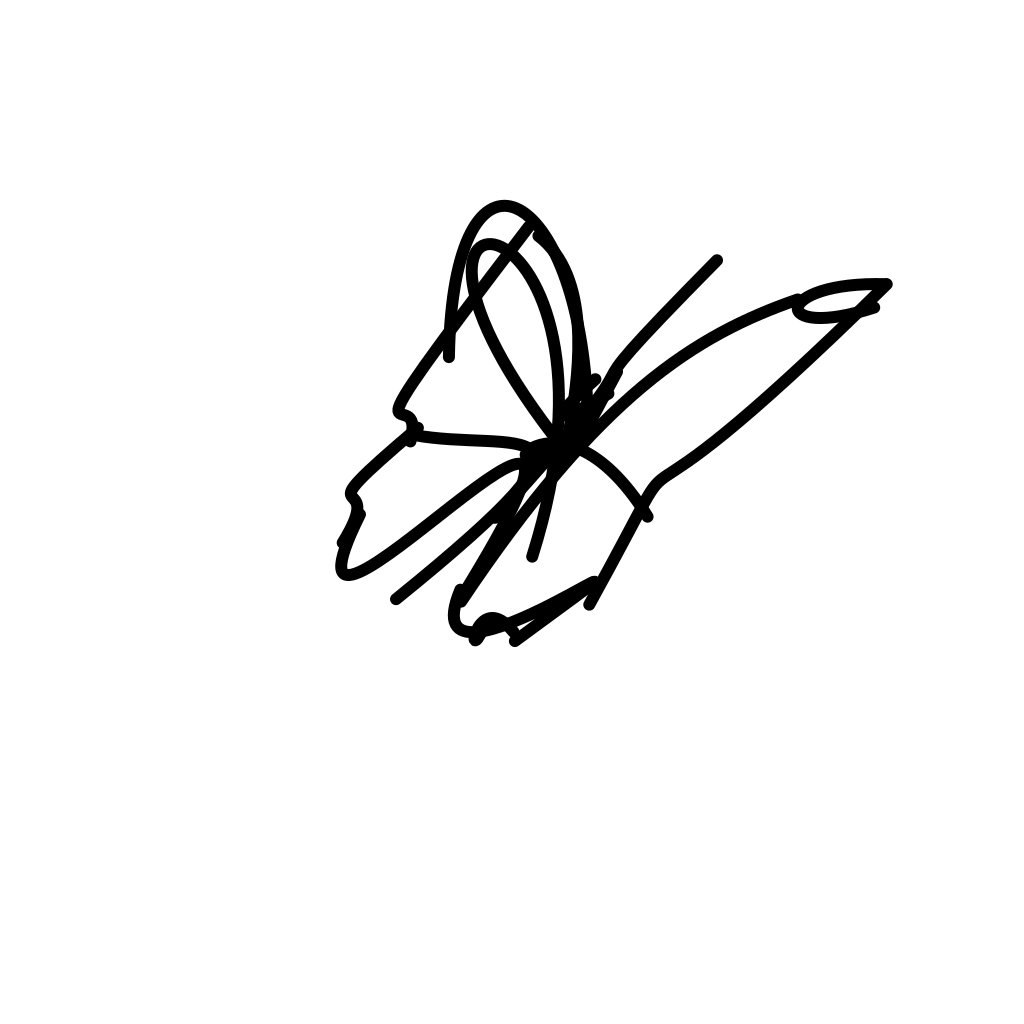}}
         & \raisebox{-0.5\height}{\includegraphics[width=.15\linewidth]{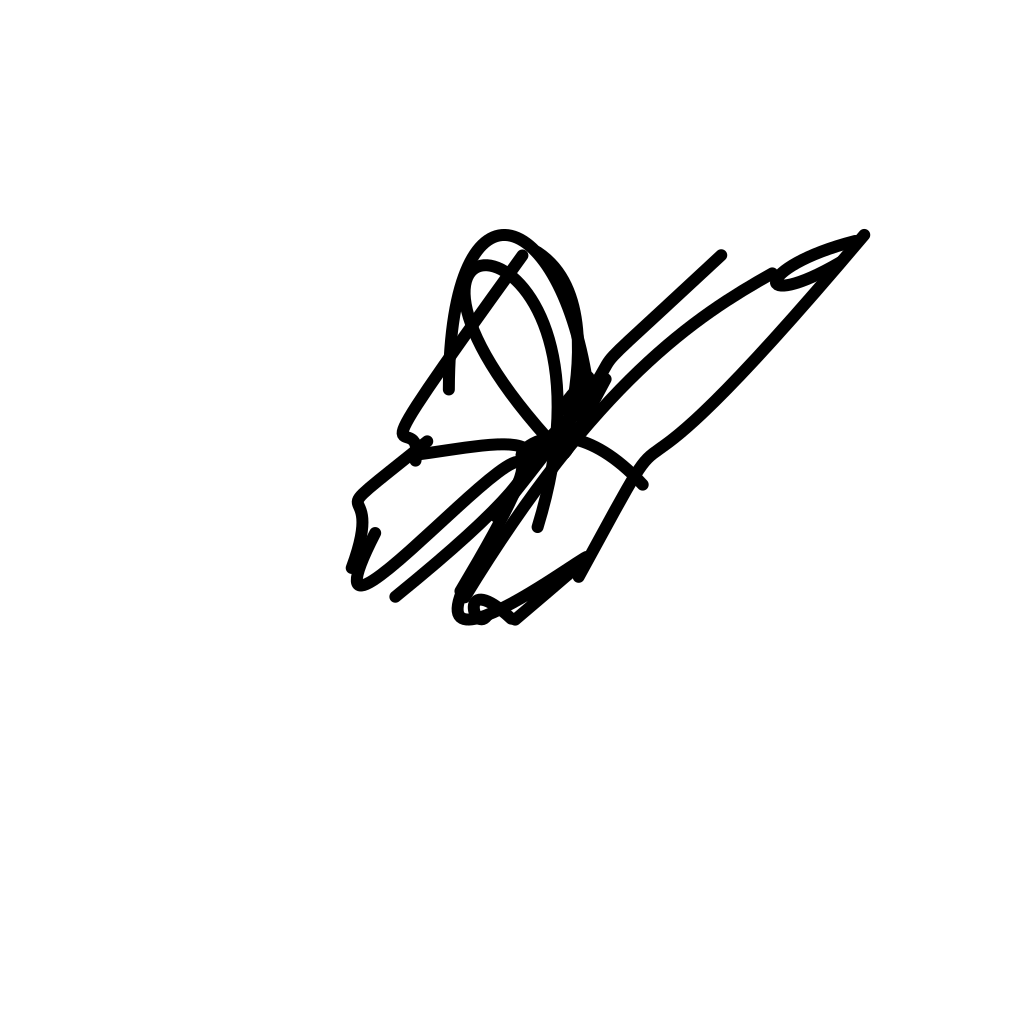}} 
         & \raisebox{-0.5\height}{\includegraphics[width=.15\linewidth]{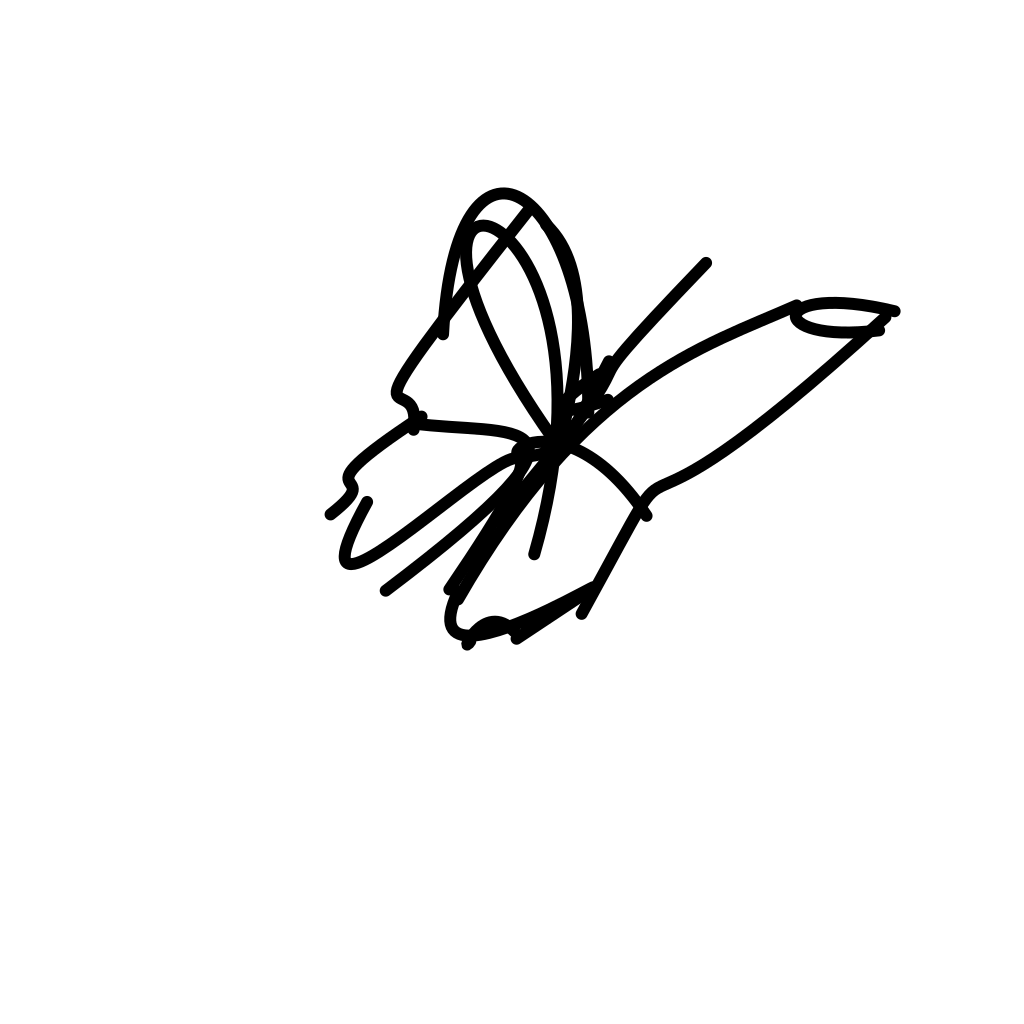}} \\
         & \multicolumn{5}{c}{\small \textcolor{blue}{Text Prompt: } "A butterfly fluttering its wings and flying gracefully."} \\
    \end{tabular}
    \caption{Qualitative results of our proposed method and generated animated sketch sequences from input text prompts.}
    \label{fig:results}
\end{figure*}

\begin{table*}
  \centering
  \footnotesize
  \caption{Comparison with state-of-the-art methods.}
  \label{tab:comparison}
  \begin{tabular}{lccc}
    \midrule
    & \textbf{Sketch-to-video consistency  $(\uparrow)$} & \textbf{Text-to-video alignment  $(\uparrow)$} \\
    \hline
    VideoCrafter1~\cite{chen2023videocrafter1} & 0.7064 & 0.0876 \\
    LiveSketch~\cite{gal2023breathing} & 0.8287 & 0.1852 \\
    Ours & \textbf{0.8561} & \textbf{0.1893} \\
    \bottomrule
  \end{tabular}
\end{table*}


\section{Experiments and results}
\subsection{Implementation details}
We use a text-to-video diffusion model~\cite{wang2023modelscope} similar to the approach in LiveSketch~\cite{gal2023breathing}, to generate the required motion in pixel space. Further, we use the generated frames to apply the SDS loss training for a timestep to find the updated control points. These updated control points are then further optimized using our learning procedure. The LA and ARAP losses are optimized using an MLP function. This MLP takes the post-LiveSketch updated control points of the current frame as input and outputs the optimized control points. We train the  MLP for 1000 iterations of the LiveSketch model. We have used t=1000 to estimate the B\'ezier curves and find their length and area. We use the values of  $\lambda_{l}$, $\lambda_{a}$ and $\lambda_{arap}$ as $0.1$, $1e-5$, and $0.1$ respectively. Further, we use similar parameters for the local and global paths given by LiveSketch~\cite{gal2023breathing}. Our method takes approximately 2 hours to generate a sequence of 24 animated sketches, each with a resolution of 256x256.

\begin{figure*}
  \centering
    \setlength{\tabcolsep}{-2pt} 
    \renewcommand{\arraystretch}{0.1} 
    \scalebox{0.9}{
    \begin{tabular}{ccccccc}
        \textbf{\small Input Sketch} & \multicolumn{5}{c}{$\overbrace{\rule{3in}{0pt}}^\text{\textbf{\small Video frames}}$} \\

         \raisebox{-0.0in}{\multirow{3}{*}{\large (a)}} 
         \multirow{3}{*} {
            \begin{tikzpicture}
                \node[inner sep=0pt] {\includegraphics[width=.18\linewidth]
                {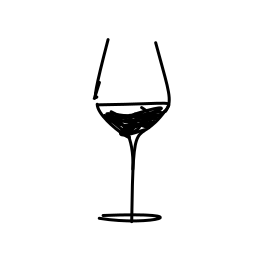}};
            \end{tikzpicture}
         } 
         & \begin{tikzpicture}
                \node {\includegraphics[width=.15\linewidth]{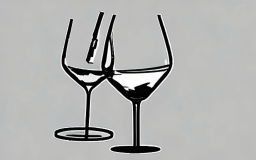}};
           \end{tikzpicture}
         & \begin{tikzpicture}
                \node {\includegraphics[width=.15\linewidth]{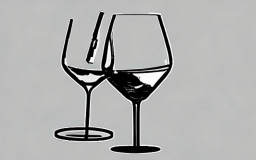}};
           \end{tikzpicture}
         & \begin{tikzpicture}
                \node {\includegraphics[width=.15\linewidth]{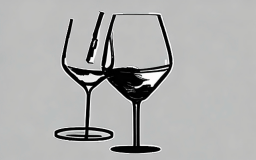}};
           \end{tikzpicture}
         & \begin{tikzpicture}
                \node {\includegraphics[width=.15\linewidth]{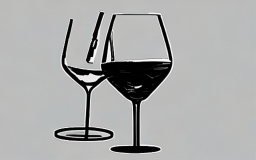}};
           \end{tikzpicture}
         & \quad \raisebox{-0.1\height}{\rotatebox{90}{\textbf{\small VideoCrafter1}}} \\[-8pt] 

        & \begin{tikzpicture}[spy using outlines={rectangle,magnification=2,size=1cm,connect spies,every spy on node/.append style={thick}}]            
            \node {\includegraphics[width=0.15\textwidth]
            {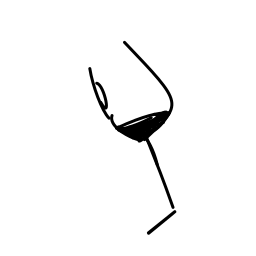}};
            \spy[color=green!50!black]  on (0.4, -0.8) in node [left] at (-0.5, -0.7);
        \end{tikzpicture}
        & \begin{tikzpicture}[spy using outlines={rectangle,magnification=2,size=1cm,connect spies,every spy on node/.append style={thick}}]       
            \node {\includegraphics[width=0.15\textwidth]{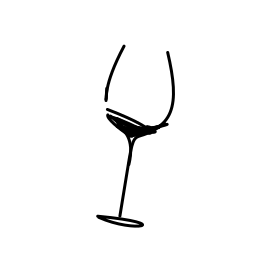}};
            \spy[color=green!50!black]  on (0.0, 0.1) in node [left] at (-0.5, -0.7);
        \end{tikzpicture}
        & \begin{tikzpicture}[spy using outlines={rectangle,magnification=2,size=1cm,connect spies,every spy on node/.append style={thick}}]          
            \node {\includegraphics[width=0.15\textwidth]{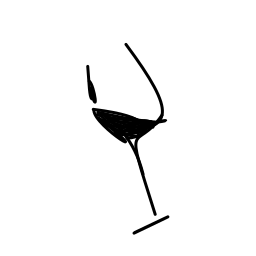}};
            \spy[color=green!50!black]  on (0.2, -0.8) in node [left] at (-0.5, -0.7);
        \end{tikzpicture}
        & \begin{tikzpicture}[spy using outlines={rectangle,magnification=2,size=1cm,connect spies,every spy on node/.append style={thick}}]      
            \node {\includegraphics[width=0.15\textwidth]{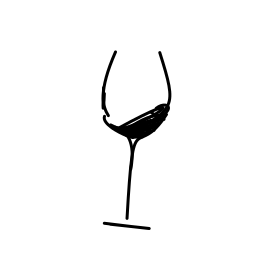}};
            \spy[color=green!50!black]  on (0.0, 0.1) in node [left] at (-0.5, -0.7);
        \end{tikzpicture}
        & \quad \raisebox{0.1\height}{\rotatebox{90}{\textbf{\small LiveSketch}}} \\[-6pt]

        & \begin{tikzpicture}[spy using outlines={rectangle,magnification=2,size=1cm,connect spies,every spy on node/.append style={thick}}]      
            \node {\includegraphics[width=0.15\linewidth]{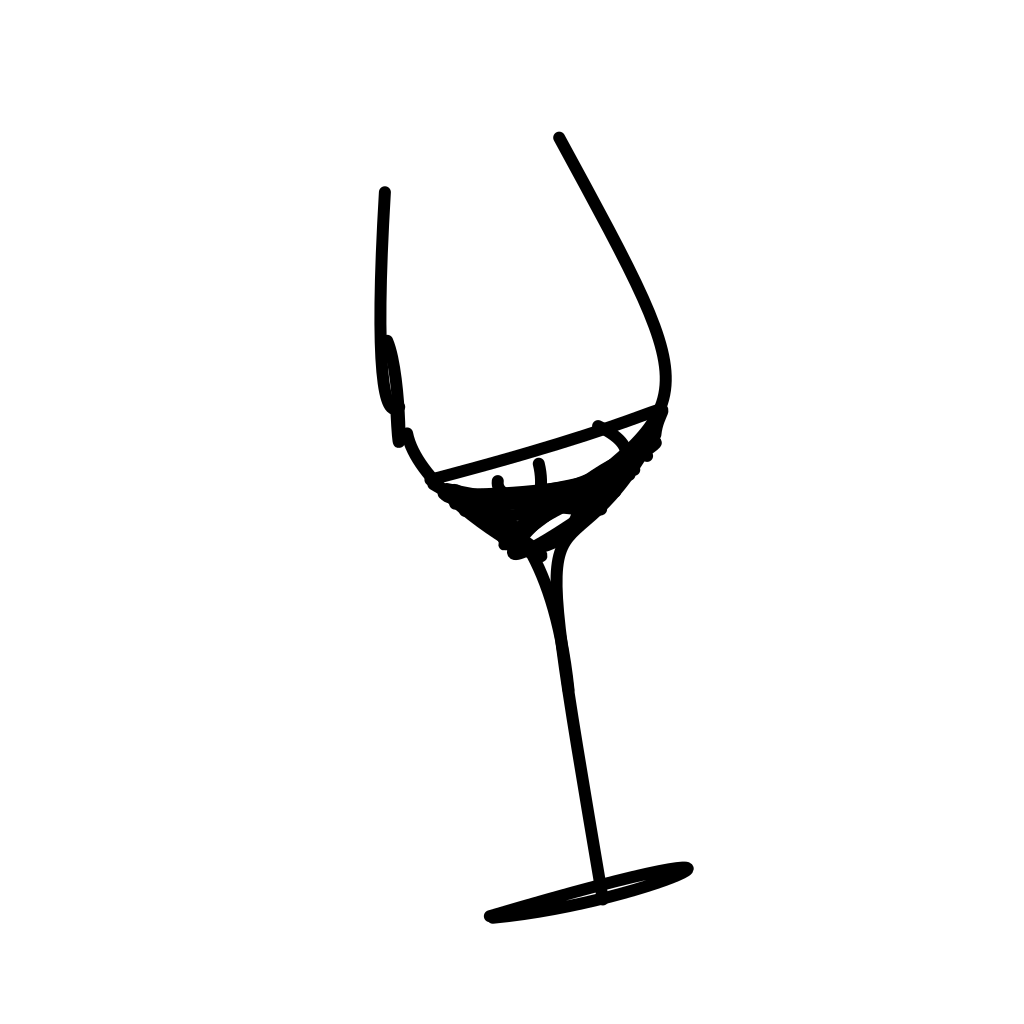}};
            \spy[color=green!50!black]  on (0.2, -0.8) in node [left] at (-0.5, -0.7);
        \end{tikzpicture}
        & \begin{tikzpicture}[spy using outlines={rectangle,magnification=2,size=1cm,connect spies,every spy on node/.append style={thick}}]      
            \node {\includegraphics[width=0.15\linewidth]{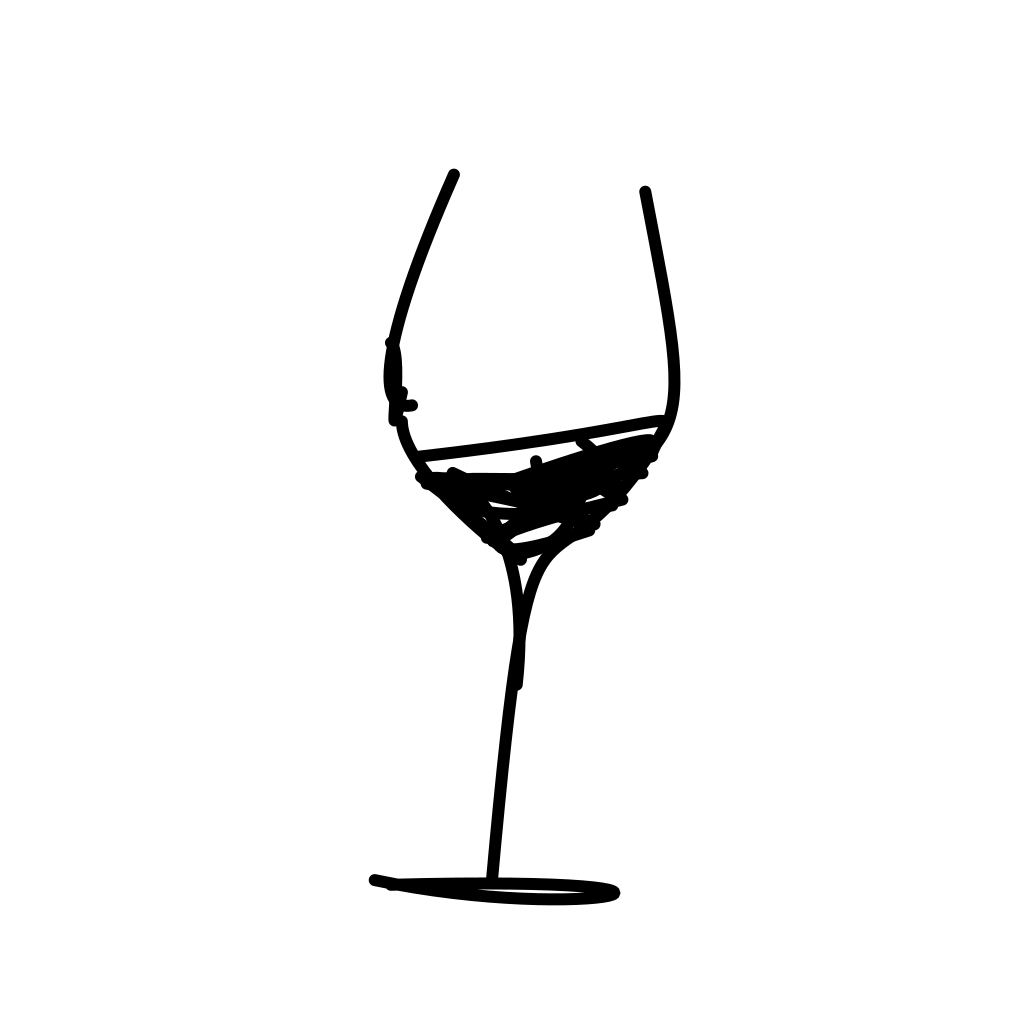}};
            \spy[color=green!50!black]  on (0.0, 0.1) in node [left] at (-0.5, -0.7);
        \end{tikzpicture}
        & \begin{tikzpicture}[spy using outlines={rectangle,magnification=2,size=1cm,connect spies,every spy on node/.append style={thick}}]      
            \node {\includegraphics[width=0.15\linewidth]{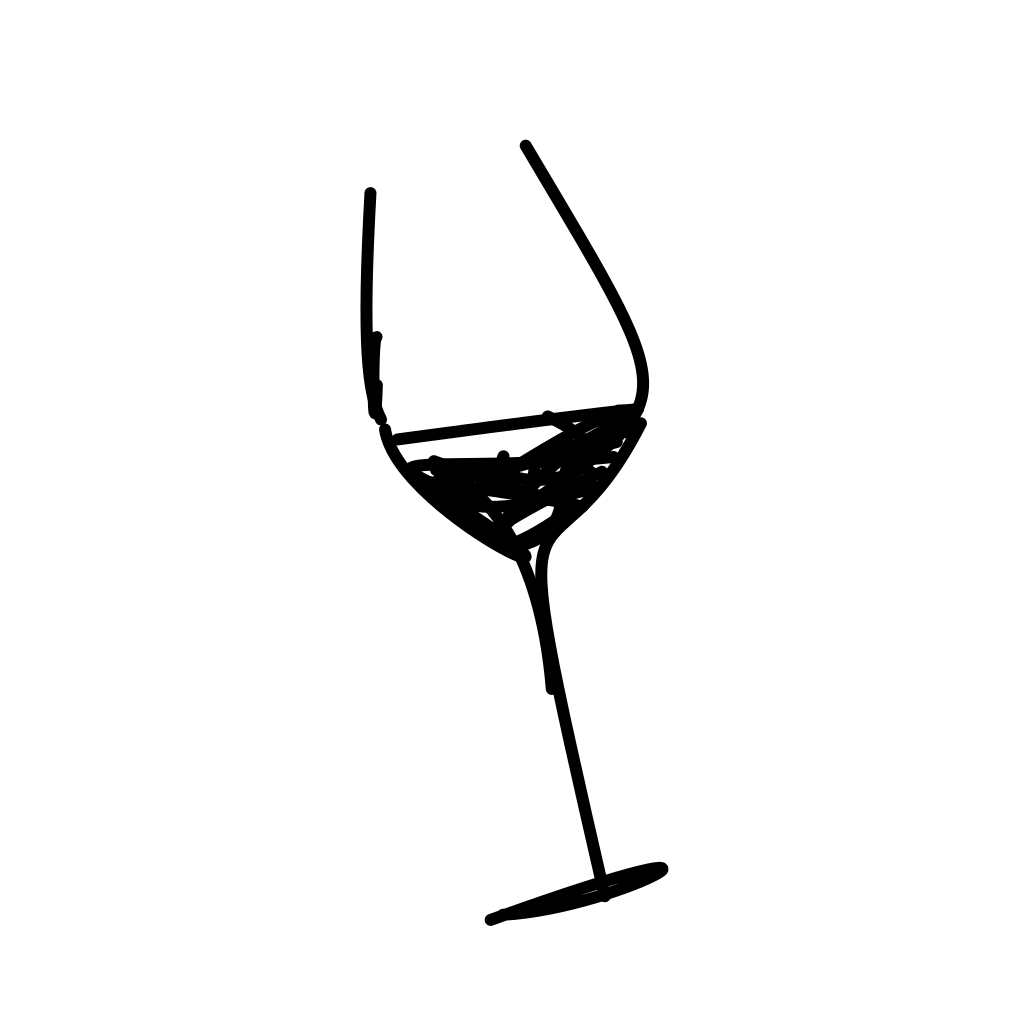}};
            \spy[color=green!50!black]  on (0.2, -0.8) in node [left] at (-0.5, -0.7);
        \end{tikzpicture}
        & \begin{tikzpicture}[spy using outlines={rectangle,magnification=2,size=1cm,connect spies,every spy on node/.append style={thick}}]      
            \node {\includegraphics[width=0.15\linewidth]{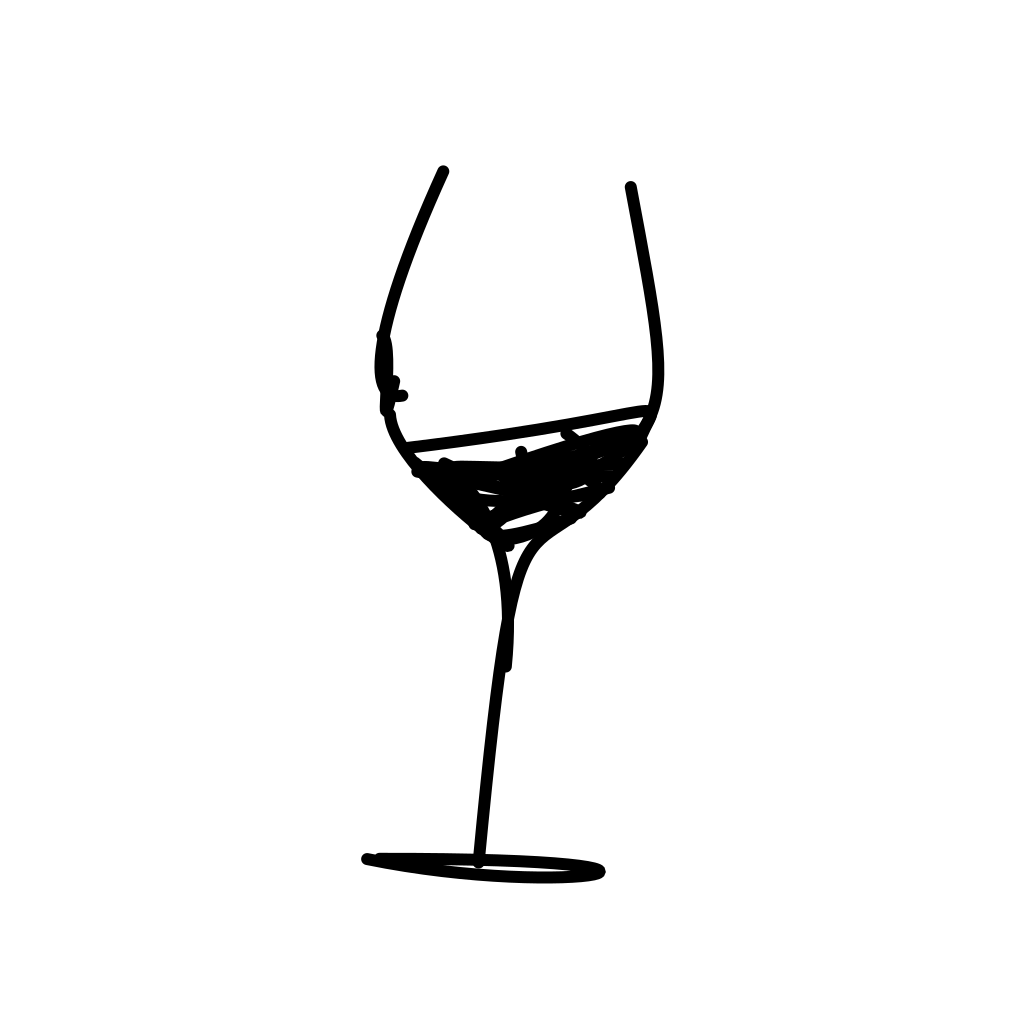}};
            \spy[color=green!50!black]  on (0.0, 0.1) in node [left] at (-0.5, -0.7);
        \end{tikzpicture}
        & \quad \raisebox{1.0\height}{\rotatebox{90}{\textbf{\small Ours}}} \\
        & \multicolumn{5}{c}{\small \textcolor{blue}{Text Prompt}: ``The wine in the wine glass sways from side to side.''} \\

         \raisebox{-0.0in}{\multirow{3}{*}{\large (b)}}
         \multirow{3}{*}{
            \begin{tikzpicture}   
                \node[inner sep=0pt] {\includegraphics[width=.2\linewidth]{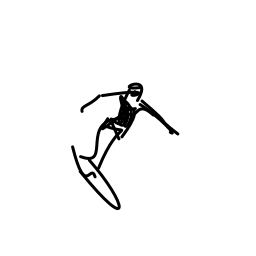}};
            \end{tikzpicture}
         }
         & \begin{tikzpicture}
                \node {\includegraphics[width=.15\linewidth]{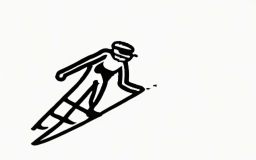}};
           \end{tikzpicture}
         & \begin{tikzpicture}
                \node {\includegraphics[width=.15\linewidth]{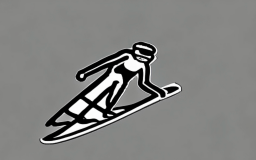}};
           \end{tikzpicture}
         & \begin{tikzpicture}
                \node {\includegraphics[width=.15\linewidth]{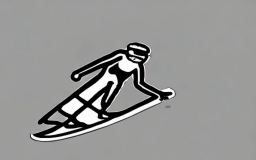}};
           \end{tikzpicture}
         & \begin{tikzpicture}
                \node {\includegraphics[width=.15\linewidth]{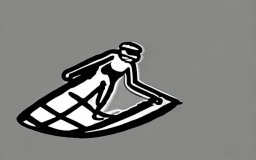}};
           \end{tikzpicture}
         & \quad \raisebox{-0.1\height}{\rotatebox{90}{\textbf{\small VideoCrafter1}}} \\[-8pt]

        & \begin{tikzpicture}[spy using outlines={rectangle,magnification=2,size=1cm,connect spies,every spy on node/.append style={thick}}]    
            \node {\includegraphics[width=0.15\textwidth]{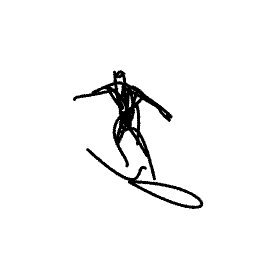}};
             \spy[color=green!50!black]  on (0.1, -0.5) in node [left] at (-0.5, -0.5);
        \end{tikzpicture}
        & \begin{tikzpicture}[spy using outlines={rectangle,magnification=2,size=1cm,connect spies,every spy on node/.append style={thick}}]    
            \node {\includegraphics[width=0.15\textwidth]{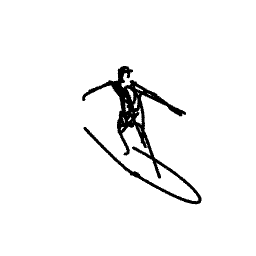}};
            \spy[color=green!50!black]  on (0.3, -0.5) in node [left] at (-0.5, -0.5);
        \end{tikzpicture}
        & \begin{tikzpicture}[spy using outlines={rectangle,magnification=2,size=1cm,connect spies,every spy on node/.append style={thick}}]    
            \node {\includegraphics[width=0.15\textwidth]{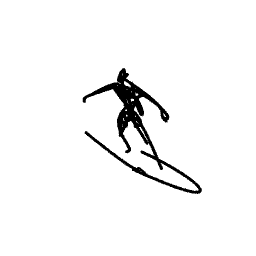}};
            \spy[color=green!50!black]  on (0.3, -0.5) in node [left] at (-0.5, -0.5);
        \end{tikzpicture}
        & \begin{tikzpicture}[spy using outlines={rectangle,magnification=2,size=1cm,connect spies,every spy on node/.append style={thick}}]    
            \node {\includegraphics[width=0.15\textwidth]{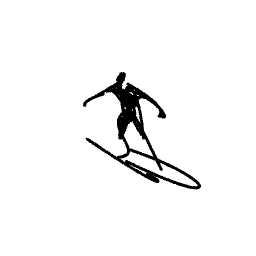}};
            \spy[color=green!50!black]  on (0.3, -0.5) in node [left] at (-0.5, -0.5);
        \end{tikzpicture}
        & \quad \raisebox{0.1\height}{\rotatebox{90}{\textbf{\small LiveSketch}}} \\[-8pt]

        & \begin{tikzpicture}[spy using outlines={rectangle,magnification=2,size=1cm,connect spies,every spy on node/.append style={thick}}]    
            \node {\includegraphics[width=0.15\linewidth]{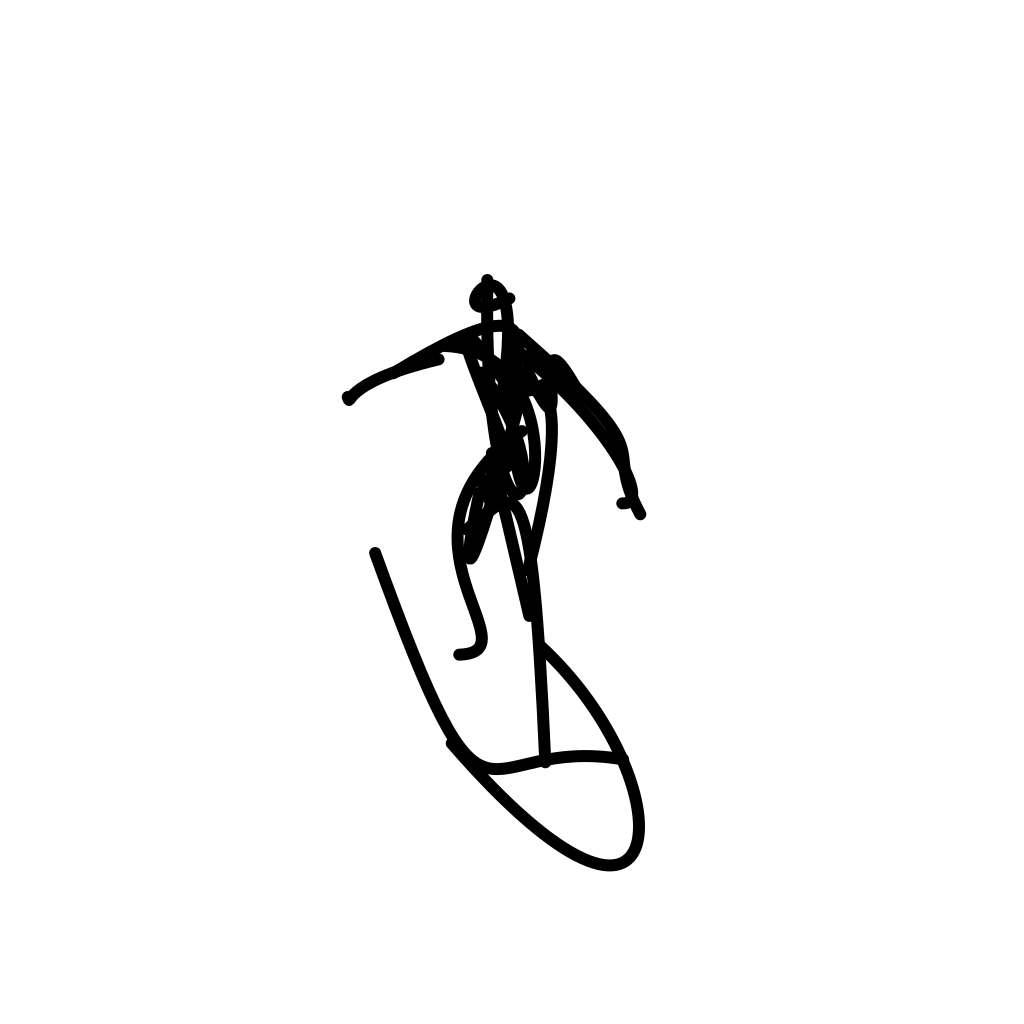}};
            \spy[color=green!50!black]  on (0.1, -0.5) in node [left] at (-0.5, -0.5);
        \end{tikzpicture}
        & \begin{tikzpicture}[spy using outlines={rectangle,magnification=2,size=1cm,connect spies,every spy on node/.append style={thick}}]    
            \node {\includegraphics[width=0.15\linewidth]{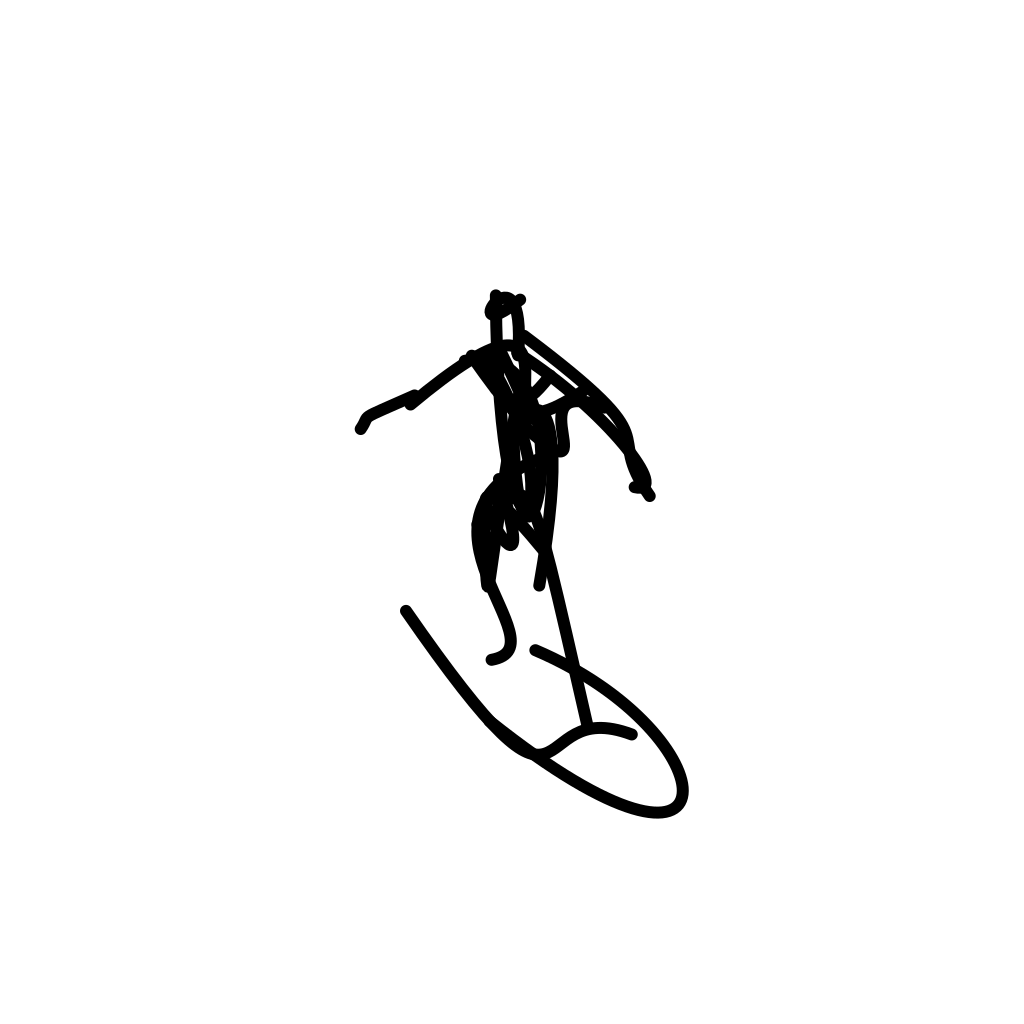}};
            \spy[color=green!50!black]  on (0.1, -0.5) in node [left] at (-0.5, -0.5);
        \end{tikzpicture}
        & \begin{tikzpicture}[spy using outlines={rectangle,magnification=2,size=1cm,connect spies,every spy on node/.append style={thick}}]    
            \node {\includegraphics[width=0.15\linewidth]{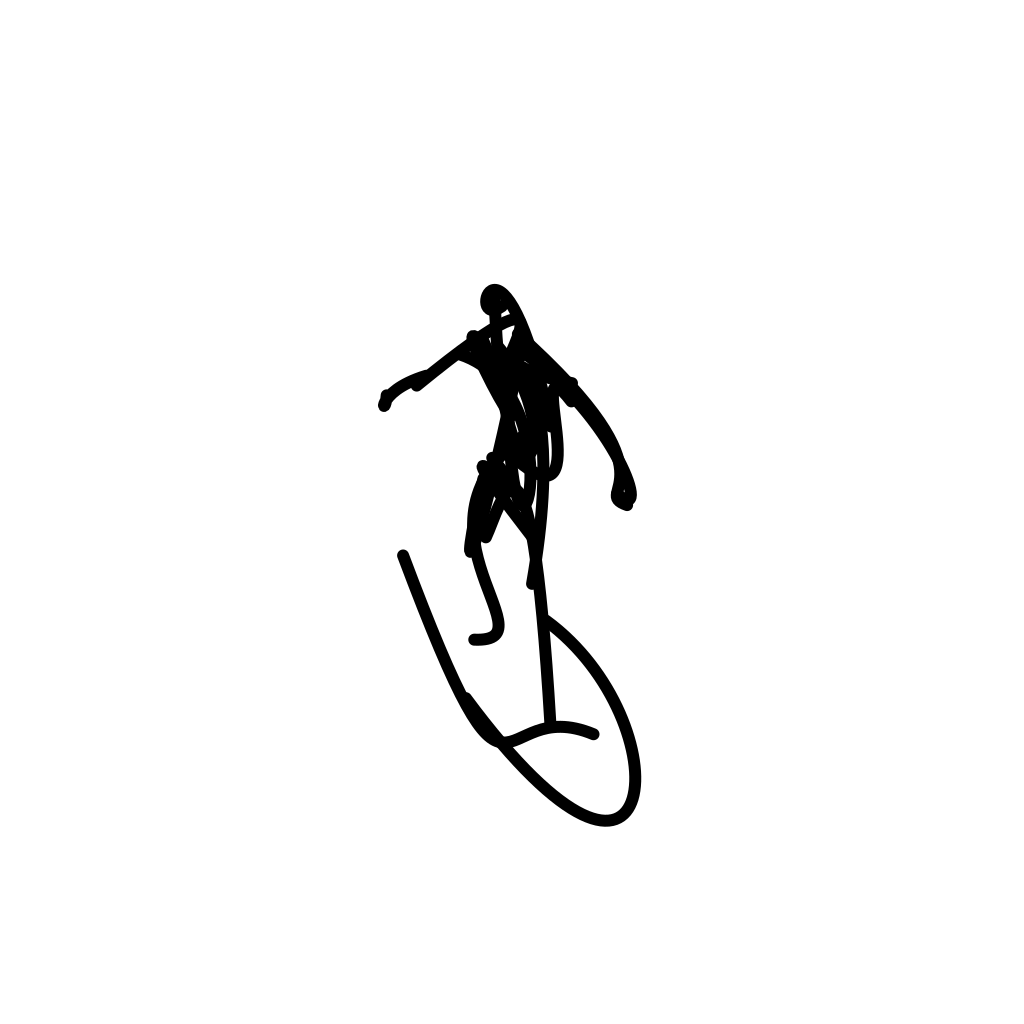}};
            \spy[color=green!50!black]  on (0.1, -0.5) in node [left] at (-0.5, -0.5);
        \end{tikzpicture}
        & \begin{tikzpicture}[spy using outlines={rectangle,magnification=2,size=1cm,connect spies,every spy on node/.append style={thick}}]    
            \node {\includegraphics[width=0.15\linewidth]{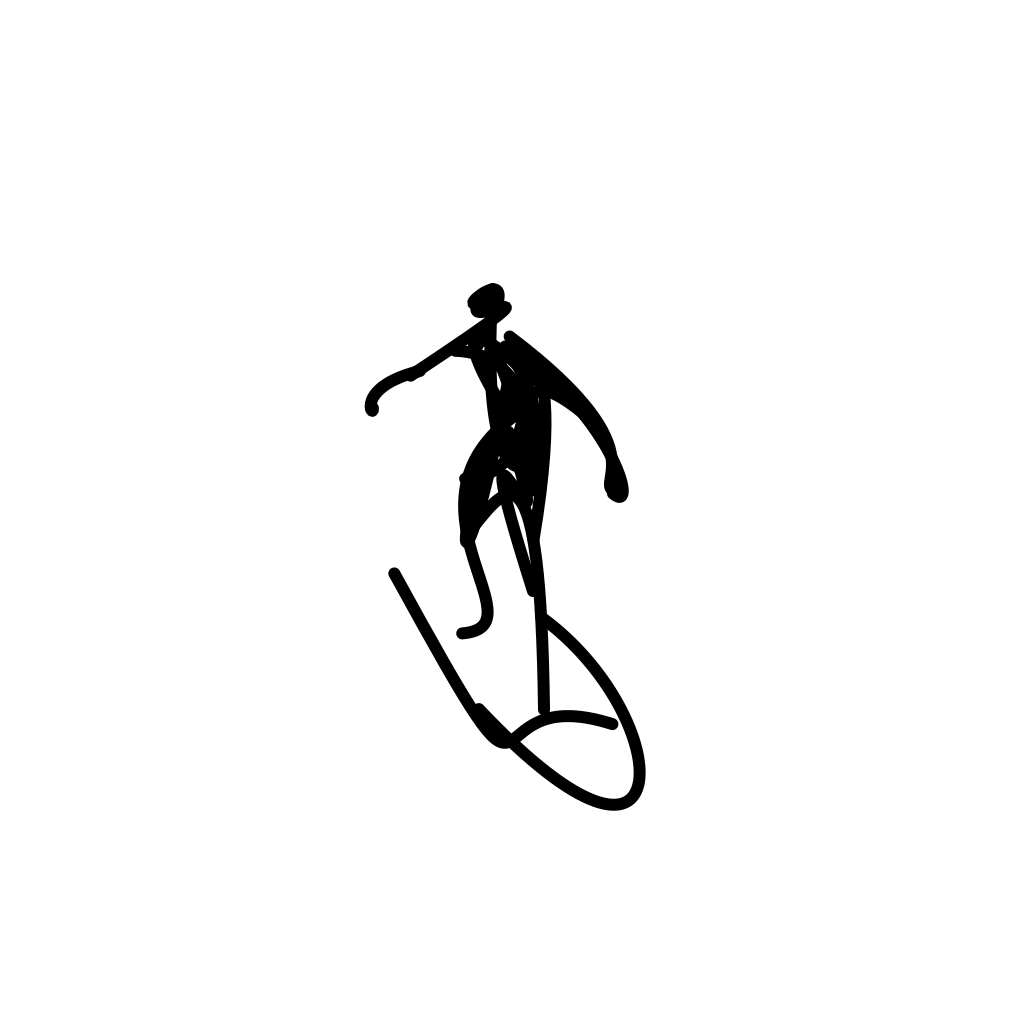}};
            \spy[color=green!50!black]  on (0.1, -0.5) in node [left] at (-0.5, -0.5);
        \end{tikzpicture}
        & \quad \raisebox{1.0\height}{\rotatebox{90}{\textbf{\small Ours}}} \\
        & \multicolumn{5}{c}{\small \textcolor{blue}{Text Prompt}: ``A surfer riding and maneuvering on waves on a surfboard.''} \\

         \raisebox{-0.0in}{\multirow{3}{*}{\large (c)}} 
         \multirow{3}{*}{
            \begin{tikzpicture}[spy using outlines={rectangle,magnification=2,size=1cm,connect spies,every spy on node/.append style={thick}}]    
                \node[inner sep=0pt] {\includegraphics[width=.18\linewidth]{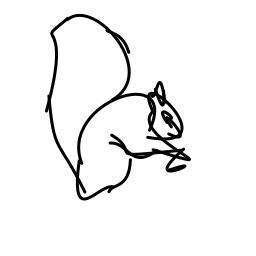}};
            \end{tikzpicture}
         }
         & \begin{tikzpicture}
                \node {\includegraphics[width=.15\linewidth]{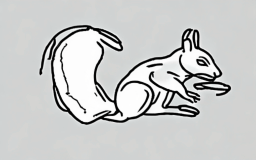}};
           \end{tikzpicture}
         & \begin{tikzpicture}
                \node {\includegraphics[width=.15\linewidth]{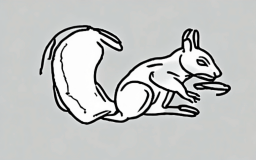}};
           \end{tikzpicture}
         & \begin{tikzpicture}
                \node {\includegraphics[width=.15\linewidth]{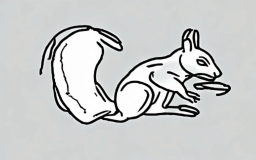}};
           \end{tikzpicture}
         & \begin{tikzpicture}
                \node {\includegraphics[width=.15\linewidth]{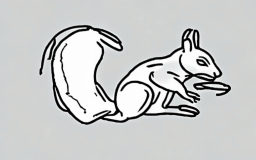}};
           \end{tikzpicture}
         & \quad \raisebox{-0.1\height}{\rotatebox{90}{\textbf{\small VideoCrafter1}}} \\[-8pt]

        & \begin{tikzpicture}[spy using outlines={rectangle,magnification=2,size=1cm,connect spies,every spy on node/.append style={thick}}]    
            \node {\includegraphics[width=0.15\textwidth]{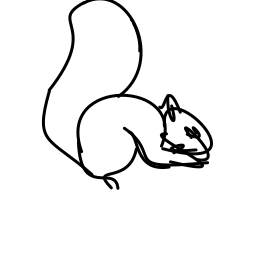}};
            \spy[color=green!50!black]  on (0.5, 0.1) in node [left] at (-0.6, -0.6);
        \end{tikzpicture}
        & \begin{tikzpicture}[spy using outlines={rectangle,magnification=2,size=1cm,connect spies,every spy on node/.append style={thick}}]    
            \node {\includegraphics[width=0.15\textwidth]{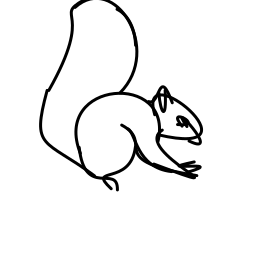}};
             \spy[color=green!50!black]  on (0.0, 0.0) in node [left] at (-0.6, -0.6);
        \end{tikzpicture}
        & \begin{tikzpicture}[spy using outlines={rectangle,magnification=2,size=1cm,connect spies,every spy on node/.append style={thick}}]    
            \node {\includegraphics[width=0.15\textwidth]{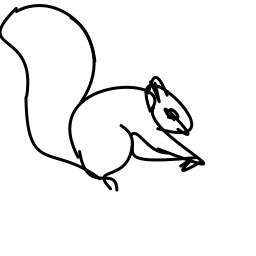}};
             \spy[color=green!50!black]  on (0.5, -0.1) in node [left] at (-0.6, -0.6);
        \end{tikzpicture}
        & \begin{tikzpicture}[spy using outlines={rectangle,magnification=2,size=1cm,connect spies,every spy on node/.append style={thick}}]    
            \node {\includegraphics[width=0.15\textwidth]{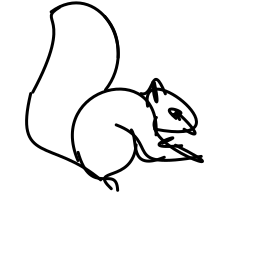}};
             \spy[color=green!50!black]  on (0.5, -0.1) in node [left] at (-0.6, -0.6);
        \end{tikzpicture}
        & \quad \raisebox{0.2\height}{\rotatebox{90}{\textbf{\small LiveSketch}}} \\[-8pt]

        & \begin{tikzpicture}[spy using outlines={rectangle,magnification=2,size=1cm,connect spies,every spy on node/.append style={thick}}]    
            \node {\includegraphics[width=0.15\linewidth]{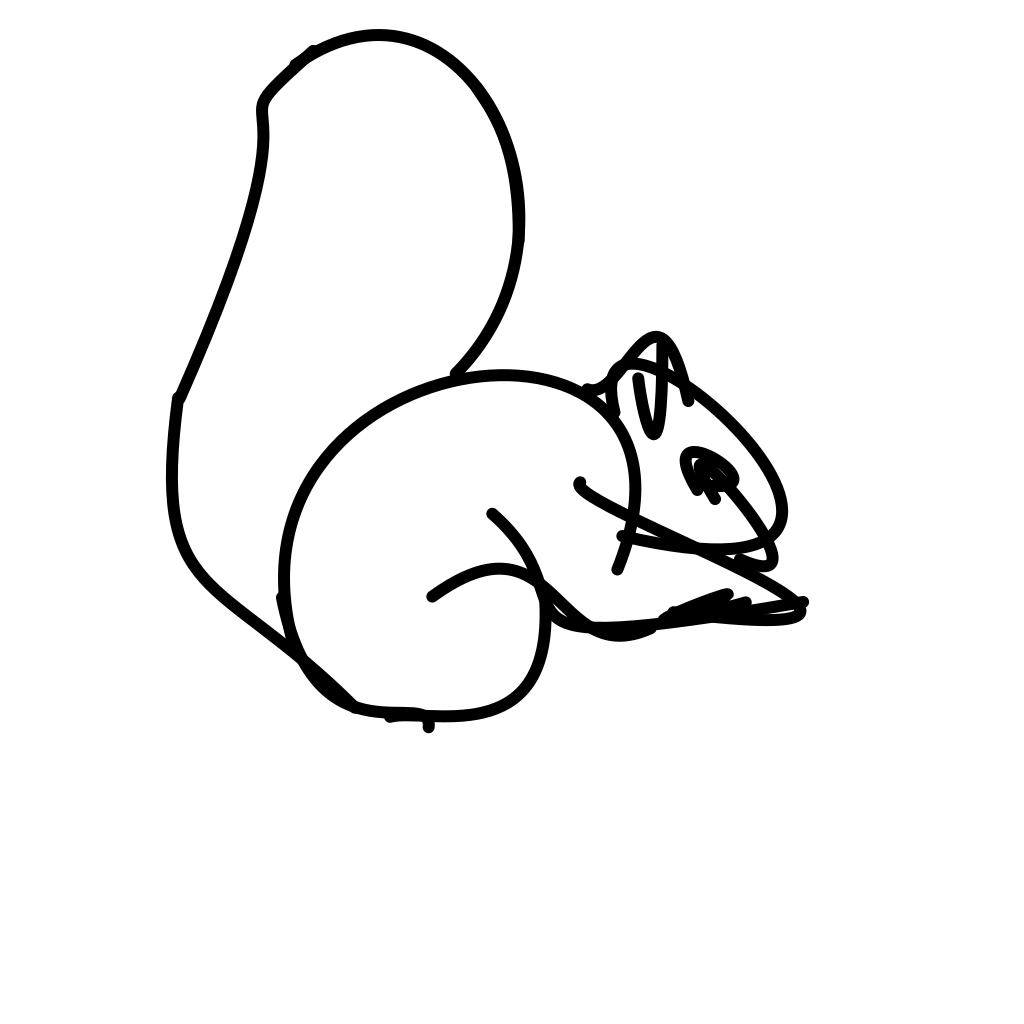}};
             \spy[color=green!50!black]  on (0.5, 0.2) in node [left] at (-0.6, -0.6);
        \end{tikzpicture}
        & \begin{tikzpicture}[spy using outlines={rectangle,magnification=2,size=1cm,connect spies,every spy on node/.append style={thick}}]    
            \node {\includegraphics[width=0.15\linewidth]{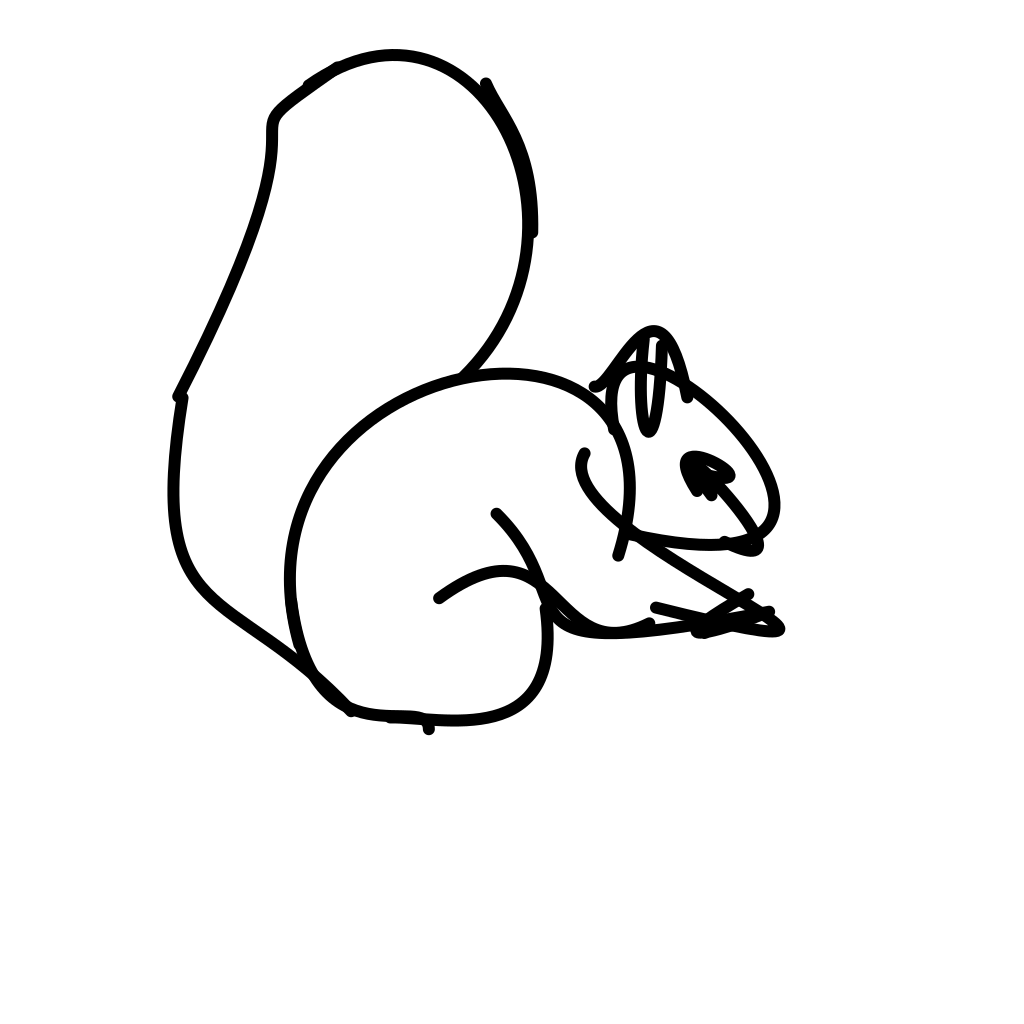}};
             \spy[color=green!50!black]  on (-0.1, 0.0) in node [left] at (-0.6, -0.6);
        \end{tikzpicture}
        & \begin{tikzpicture}[spy using outlines={rectangle,magnification=2,size=1cm,connect spies,every spy on node/.append style={thick}}]    
            \node {\includegraphics[width=0.15\linewidth]{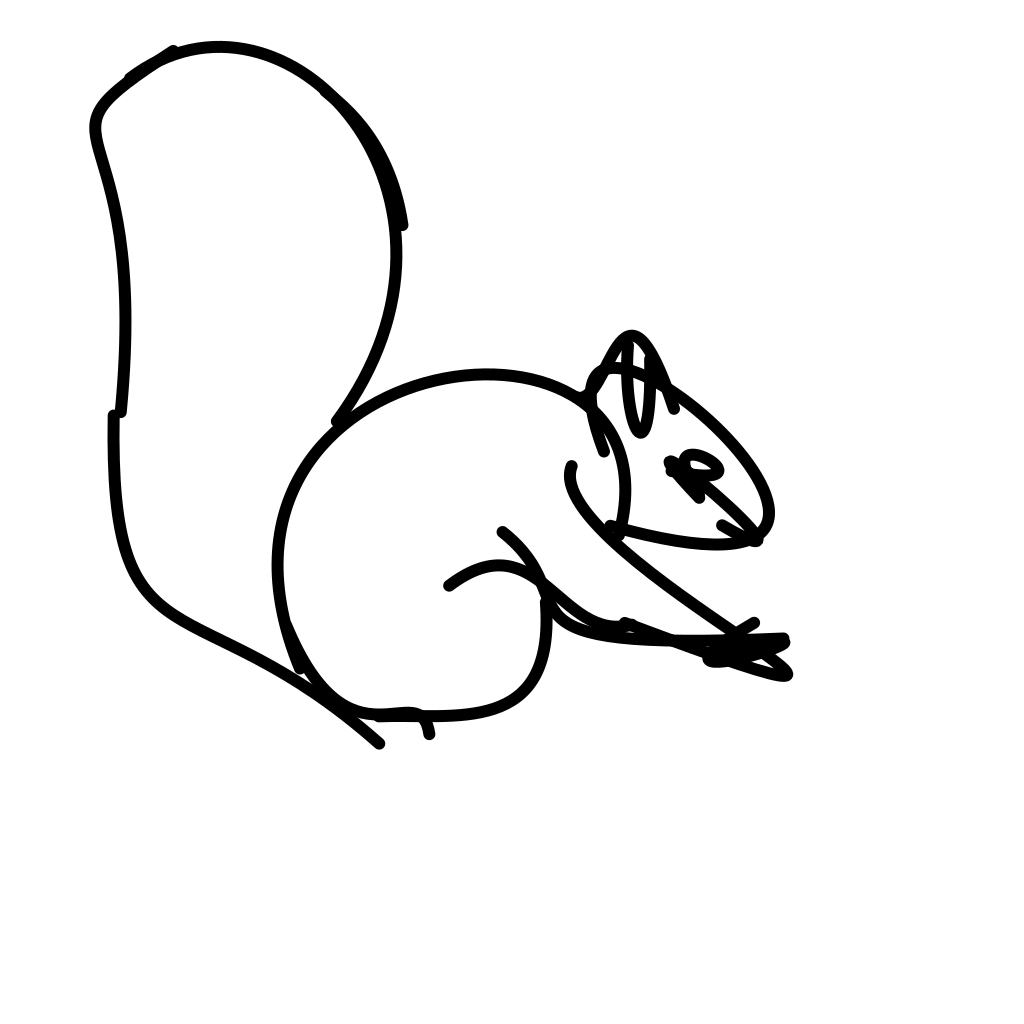}};
             \spy[color=green!50!black]  on (0.5, -0.1) in node [left] at (-0.6, -0.6);
        \end{tikzpicture}
        & \begin{tikzpicture}[spy using outlines={rectangle,magnification=2,size=1cm,connect spies,every spy on node/.append style={thick}}]    
            \node {\includegraphics[width=0.15\linewidth]{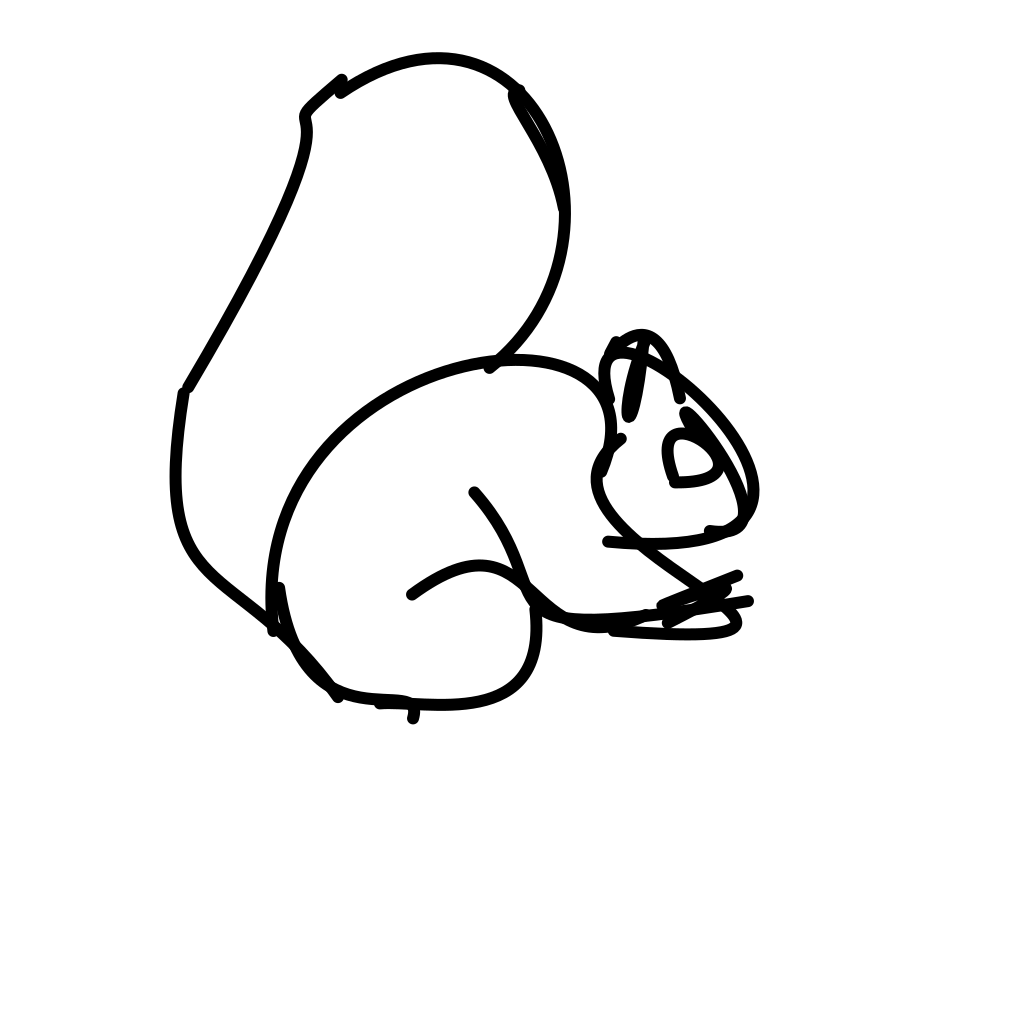}};
             \spy[color=green!50!black]  on (0.5, -0.1) in node [left] at (-0.6, -0.6);
        \end{tikzpicture}
        & \quad \raisebox{1.0\height}{\rotatebox{90}{\textbf{\small Ours}}} \\
        & \multicolumn{5}{c}{\small \textcolor{blue}{Text Prompt}: ``The squirrel uses its dexterous front paws to hold and} \\
        & \multicolumn{5}{c}{\small manipulate nuts, displaying meticulous and deliberate motions while eating.''} \\
    \end{tabular}
    }
    \caption{Comparison with state-of-the-art methods (VideoCrafter1~\cite{chen2023videocrafter1} and LiveSketch~\cite{gal2023breathing}) . In the above figure, the base of the wine glass is distorted in the previous methods, and in the surfer base, the original shape is missing compared to ours. The squirrel tails and body shape contain the original topology in our method.}
    \label{fig:comparison}
\end{figure*}

\subsection{Results and Comparison}
\subsubsection{Quantitative evaluation}
We compare our approach with previous baseline methods VideoCrafter1~\cite{chen2023videocrafter1}, and LiveSketch~\cite{gal2023breathing}. We use sketch-to-video consistency and Text-to-video alignment as evaluation matrices similar to LiveSketch~\cite{gal2023breathing} that use CLIP~\cite{radford2021learning} to estimate sketch-to-video consistency and X-CLIP~\cite{ni2022expanding} for text-to-video alignment. We used 20 unique sketch samples and text for the quantitative evaluation. VideoCrafter1~\cite{chen2023videocrafter1} is an image-to-video generation model and the conditions on image and text prompts. Table~\ref{tab:comparison} depicts that our method outperforms the quantitative analysis compared to previous methods. We maintain the text-to-video alignment similar to LiveSketch~\cite{gal2023breathing}, but the sketch-to-video consistency performance is superior to our method.

\subsubsection{Qualitative evaluation}
In the qualitative comparison, we measure the sketch-to-video consistency and Text-to-video alignment in the previous state-of-the-art methods compared to our approach. In the streamline, we further estimate the improvements such as temporal consistency and shape preservation (see Figure~\ref{fig:results}) . Sketch-to-video consistency describes the temporal consistency of the generated sketch sequences. Figure~\ref{fig:comparison} shows that the bottom of the wine glass and squirrel is temporally consistent in all the frames compared to VideoCrafter1~\cite{chen2023videocrafter1}, and LiveSketch~\cite{gal2023breathing}. 
On the other hand, we observe that the surfer and squirrel examples preserve the original shape during animation. The mesh-based rigidity loss helps to produce a smooth deformation compared to the baseline methods. Our method maintains temporal consistency, preserves shape during animation, and outperforms baseline methods.

\begin{figure*}
  \centering
    \setlength{\tabcolsep}{-1pt} 
    \scalebox{1.3}{
    \begin{tabular}{ccccccc}
        \textbf{\scriptsize Input Sketch} & \multicolumn{5}{c}{$\overbrace{\rule{3in}{0pt}}^\text{\textbf{\scriptsize Video frames}}$} \\
         \raisebox{-0.5in}{\multirow{3}{*}{\scriptsize }} \multirow{3}{*} {\includegraphics[width=.15\linewidth]{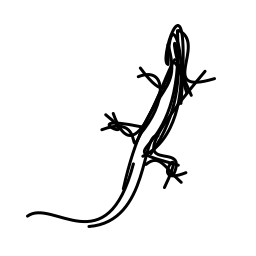}}
         & \raisebox{-0.5\height}{\includegraphics[width=.125\linewidth]{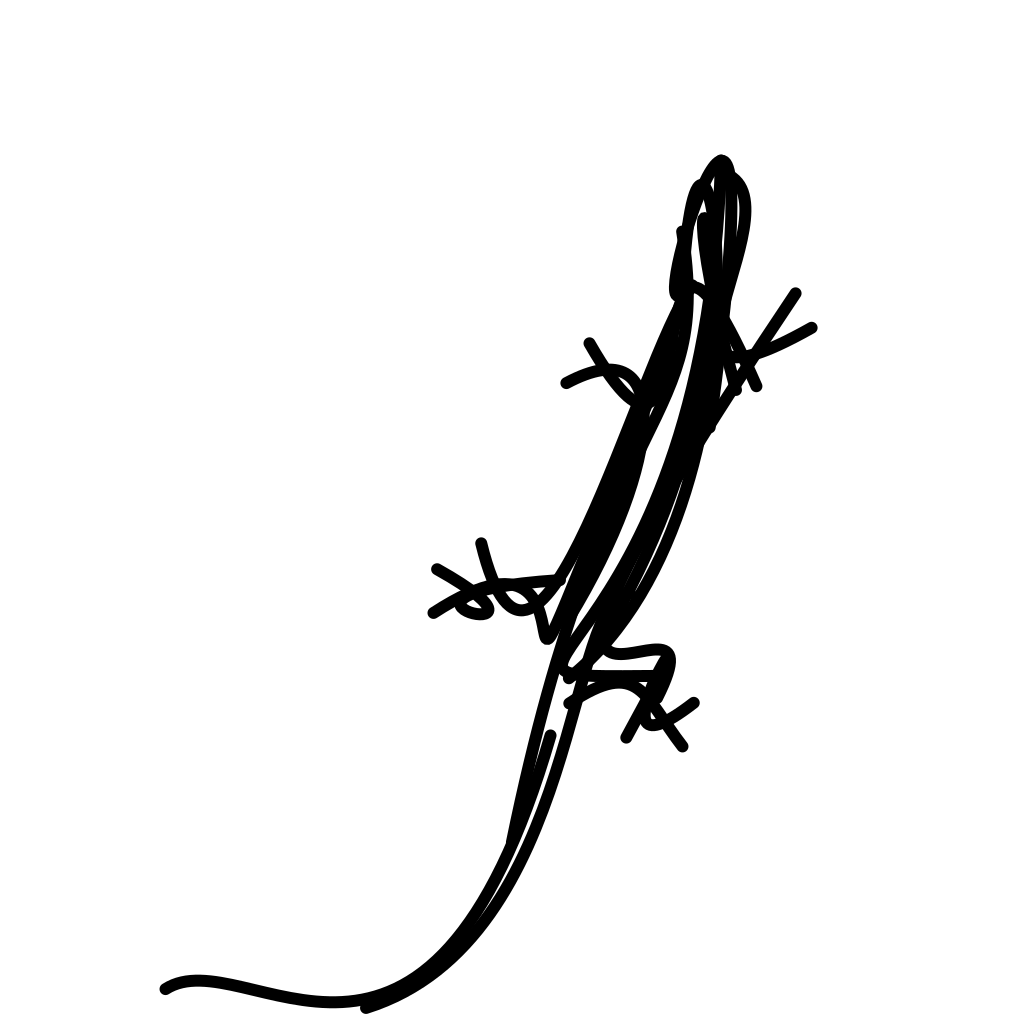}}
         & \raisebox{-0.5\height}{\includegraphics[width=.125\linewidth]{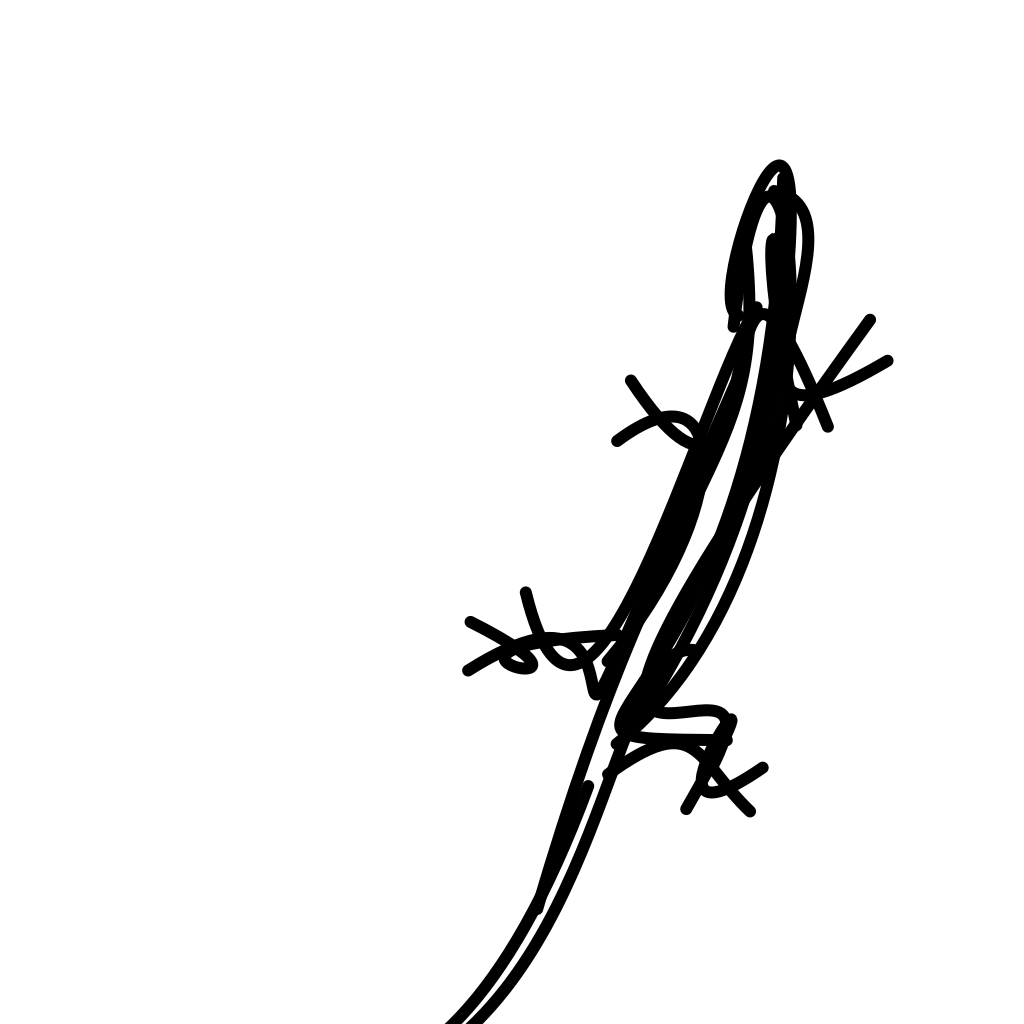}}
         & \raisebox{-0.5\height}{\includegraphics[width=.125\linewidth]{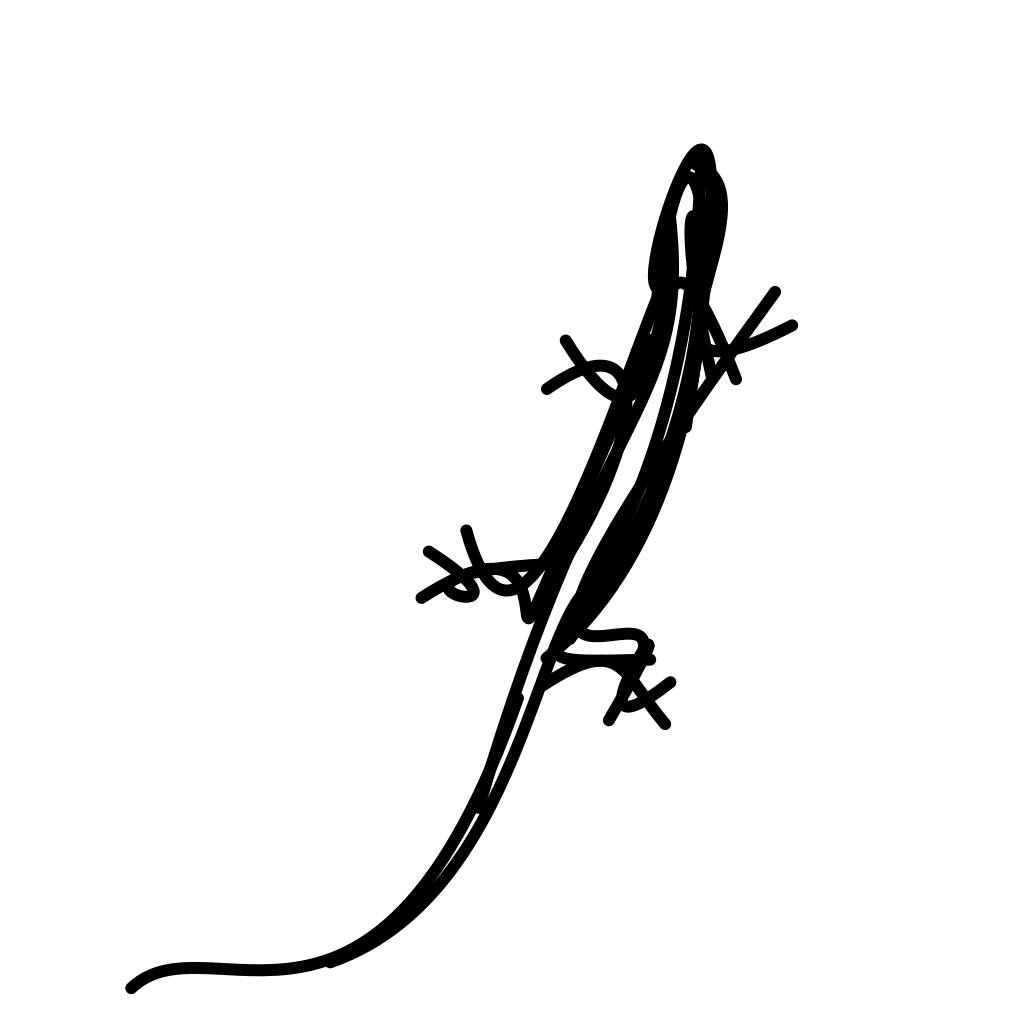}} 
         & \raisebox{-0.5\height}{\includegraphics[width=.125\linewidth]{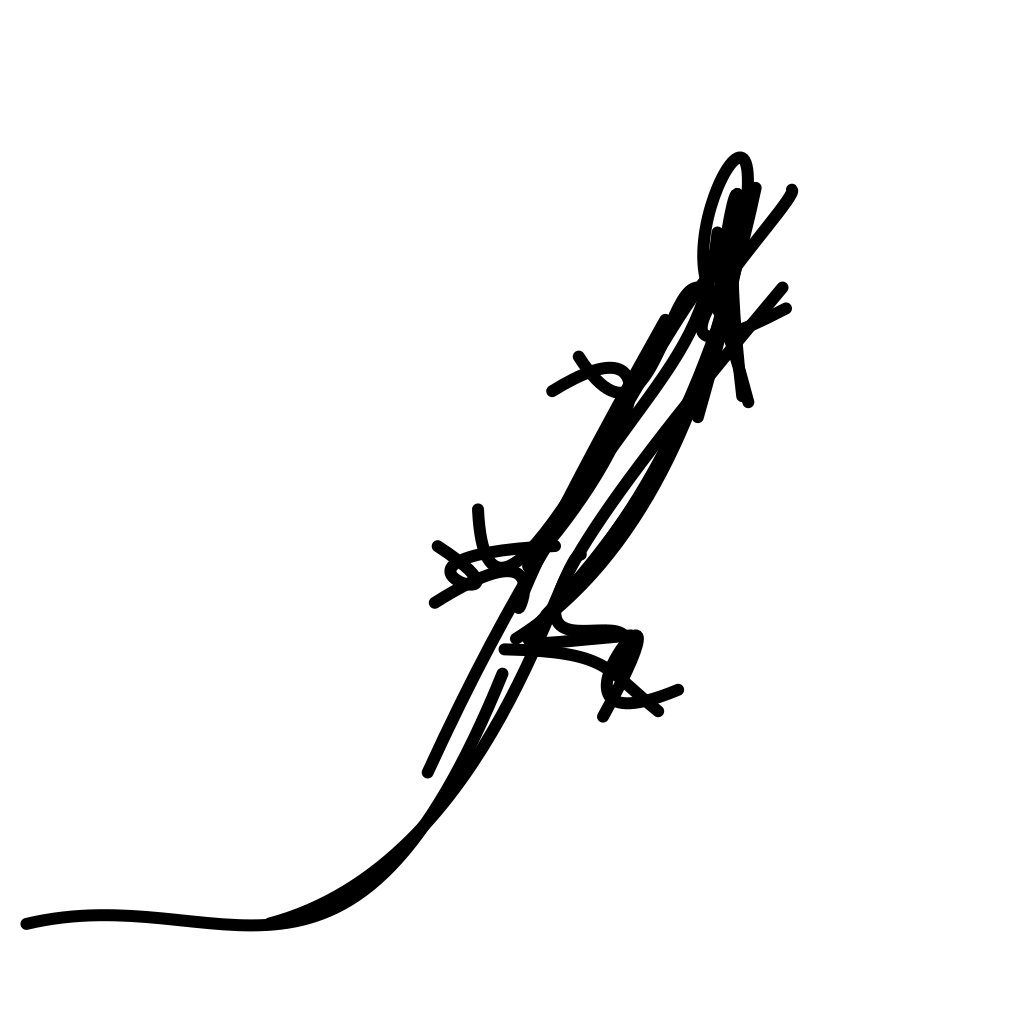}} 
         & \quad \raisebox{-0.5\height}{\rotatebox{90}{\textbf{\scriptsize w/o LA reg.}}} \\
         & \raisebox{-0.5\height}{\includegraphics[width=.125\linewidth]{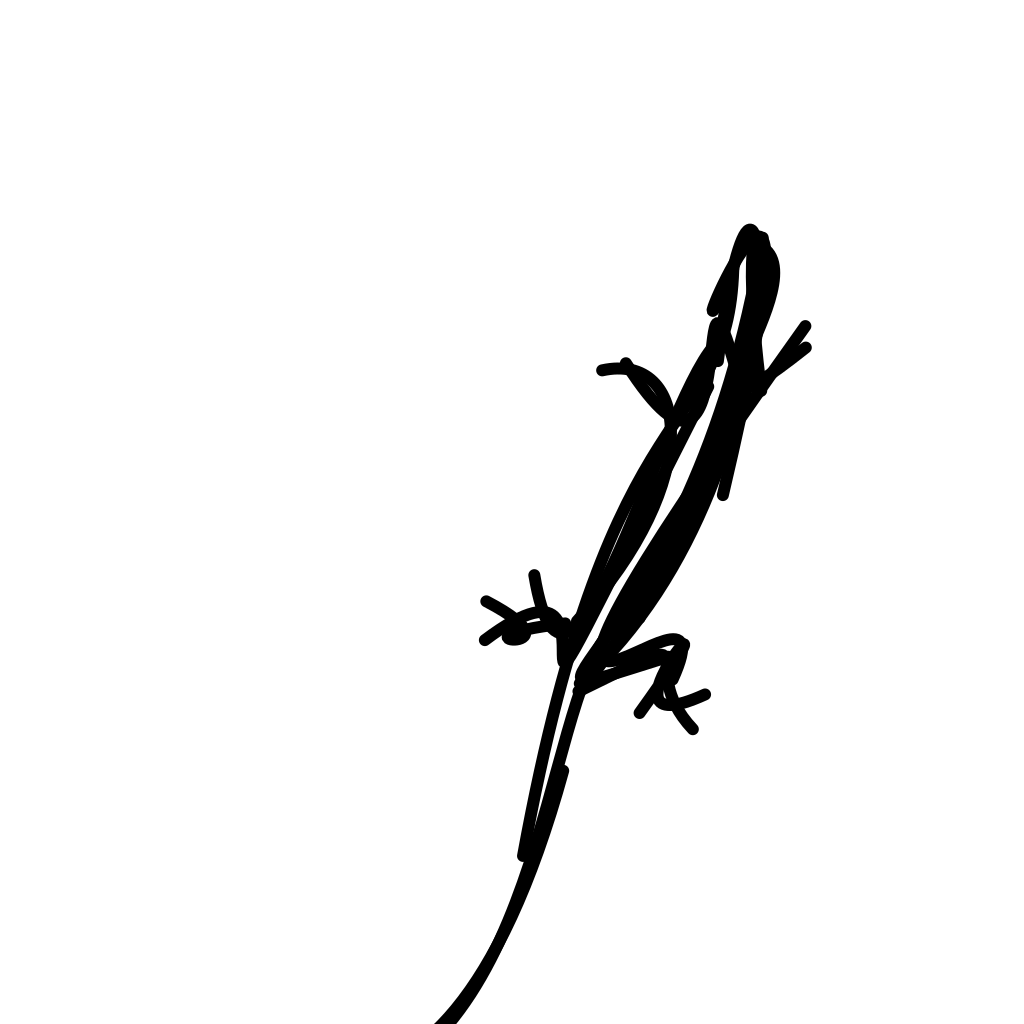}}
         & \raisebox{-0.5\height}{\includegraphics[width=.125\linewidth]{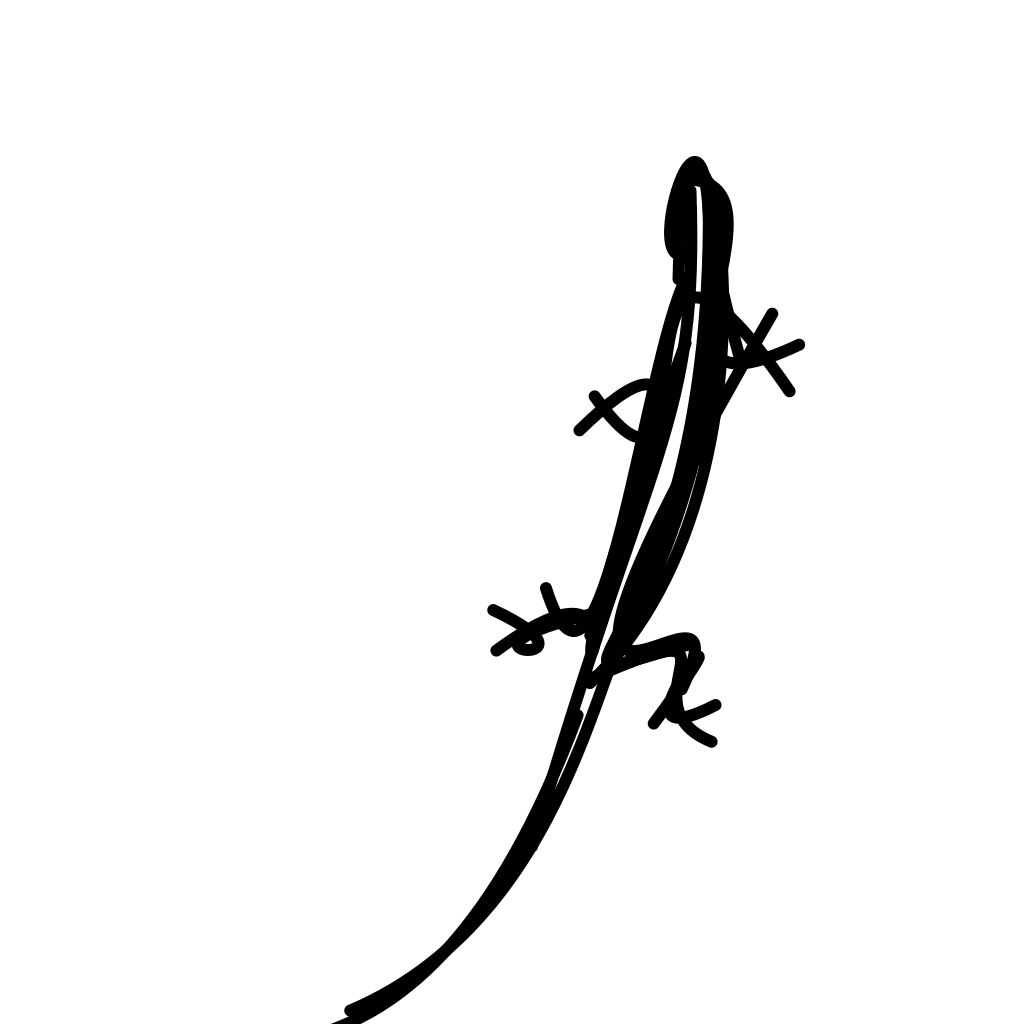}}
         & \raisebox{-0.5\height}{\includegraphics[width=.125\linewidth]{figures/ablation/lizard/wa_2.png}} 
         & \raisebox{-0.5\height}{\includegraphics[width=.125\linewidth]{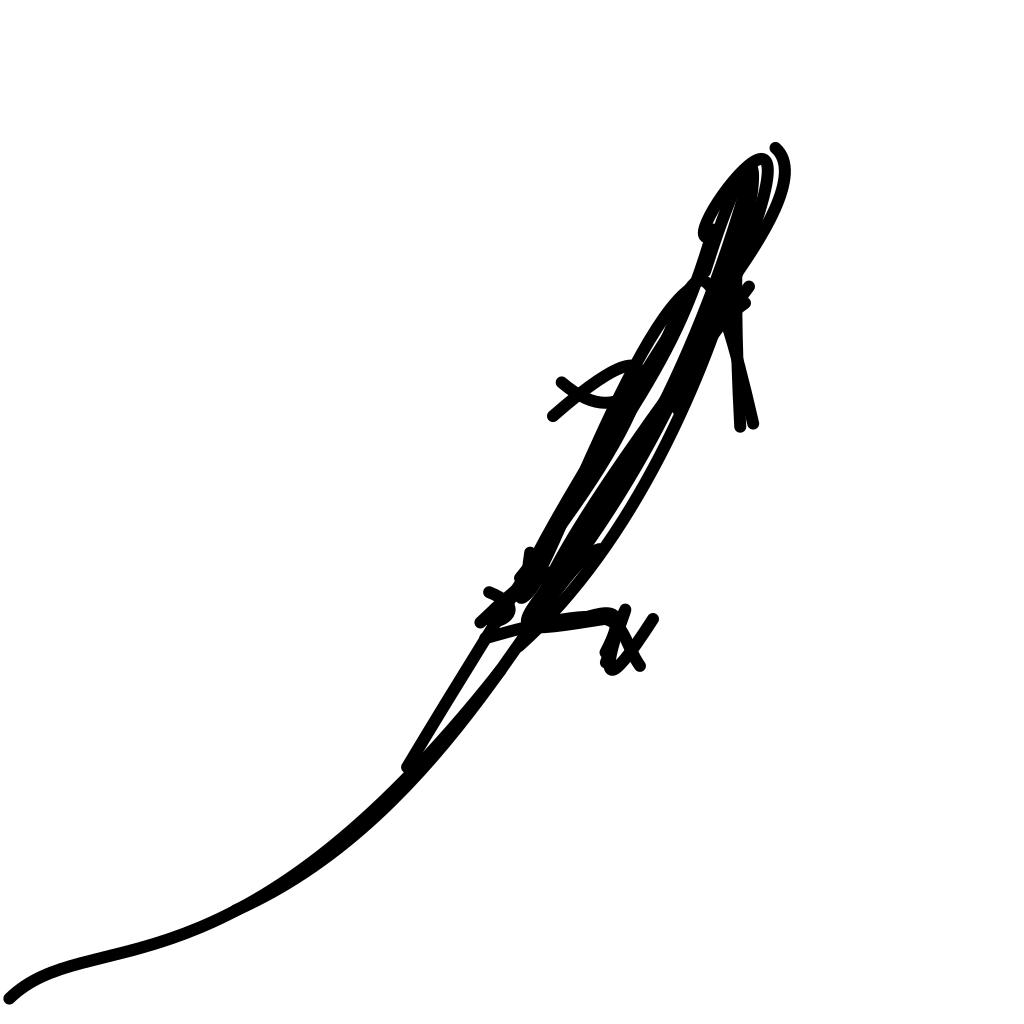}} 
         & \quad \raisebox{-0.5\height}{\rotatebox{90}{\textbf{\scriptsize w/o arap loss}}} \\
         & \raisebox{-0.5\height}{\includegraphics[width=.125\linewidth]{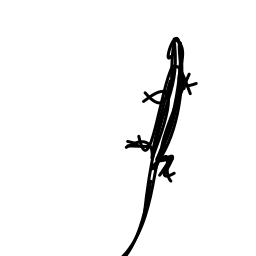}}
         & \raisebox{-0.5\height}{\includegraphics[width=.125\linewidth]{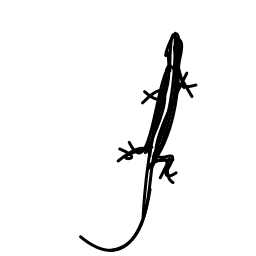}}
         & \raisebox{-0.5\height}{\includegraphics[width=.125\linewidth]{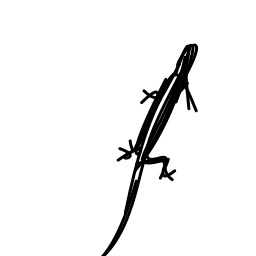}} 
         & \raisebox{-0.5\height}{\includegraphics[width=.125\linewidth]{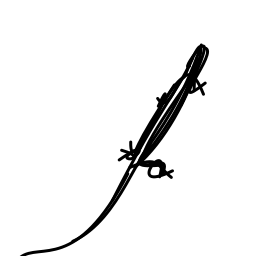}} 
         & \quad \raisebox{-0.5\height}{\rotatebox{90}{\textbf{\scriptsize Ours}}} \\
         & \multicolumn{5}{c}{\scriptsize \textcolor{blue}{Text Prompt}: ``The lizard moves with a sinuous, undulating motion, gliding} \\
         & \multicolumn{5}{c}{\scriptsize smoothly over surfaces using its agile limbs and tail for balance and propulsion.''} 
    \end{tabular}
    }
    \caption{Ablation study on different setting such as w/o LA regularization, w/o shape preserving ARAP, and our full method.}
    \label{fig:ablation}
\end{figure*}

\subsection{Ablation Study}

\subsubsection{With and without Regularization}
We evaluate our method without LA regularization and observe that it fails to maintain temporal consistency. LA regularization helps to address the issue of drastic changes in stroke. In Figure~\ref{fig:ablation}, the lizard's tail and legs move rapidly, and the stroke length varies excessively. In our proposed method, we can see the smooth motion and nominal change in stroke length. Table~\ref{tab:ablation} shows the quantitative results w/o the LA regularizer and our method with the LA regularization.

\subsubsection{With and without Shape-preserving ARAP}
We evaluate the method without shape preservation and observe shape distortion during animation. The animated sketch shows distortion when local motion increases, as topology is not preserved in the animated sketch video. In Figure~\ref{fig:ablation}, the lizard body distorts during the motion, compared to our method with shape-preserving arap loss. 
Quantitatively (see Table~\ref{tab:ablation}), the performance without shape preservation is comparable, but our complete method gives better results.

\begin{table*}
  \centering
  \footnotesize
  \caption{Ablation results—the quantitative evaluation on w/o LA regularization, w/o shape preserving ARAP, and our proposed method.}
  \label{tab:ablation}
  \begin{tabular}{lcc}
    \toprule
    & \textbf{Sketch-to-video consistency $(\uparrow)$} & \textbf{Text-to-video alignment $(\uparrow)$} \\
    \midrule
    W/o LA reg. & 0.8306 & 0.1864 \\
    W/o Shape-preserving ARAP & 0.8489 & 0.1891 \\
    Ours & \textbf{0.8561} & \textbf{0.1893} \\
    \bottomrule
  \end{tabular}
\end{table*}

\begin{figure}[!ht]
\centering
\setlength{\tabcolsep}{2pt}
\scalebox{0.95}{
\begin{tabular}{ccc}
\begin{tikzpicture}[spy using outlines={rectangle,magnification=0,size=0cm,connect spies,every spy on node/.append style={thick}}]
    \node {\includegraphics[height=0.2\textwidth]{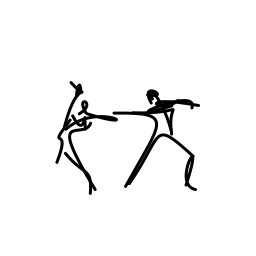}};
    \node[text width=3.5cm] at (0, -2) {\small
 \textcolor{blue}{Text Prompt}: ``The two dancers are passionately dancing the Cha-Cha, their bodies moving in sync with the infectious Latin rhythm.''};
    \node[] at (0, 1.25) {$\textbf{Input Sketch}$};
\end{tikzpicture} &
\begin{tikzpicture}[spy using outlines={rectangle,magnification=2,size=1.7cm,connect spies,every spy on node/.append style={thick}}]
    \node {\includegraphics[height=0.25\textwidth]{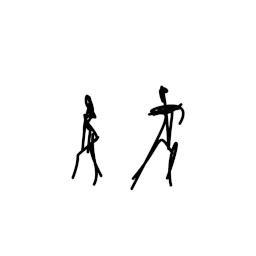}};
    \node[] at (0, 2.475) {$\textbf{Ours}$};
\end{tikzpicture} \\[-20pt]

\begin{tikzpicture}[spy using outlines={rectangle,magnification=4,size=1.7cm,connect spies,every spy on node/.append style={thick}}]
    \node {\includegraphics[height=0.2\textwidth]{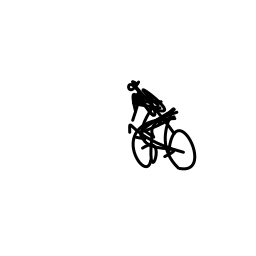}};
    \node[text width=3.5cm] at (0, -2) {\small
     \textcolor{blue}{Text Prompt}: ``The biker is pedaling, each leg pumping up and down as the wheels of the bicycle spin rapidly, propelling them forward.''};
\end{tikzpicture} &
\begin{tikzpicture}[spy using outlines={rectangle,magnification=2,size=1.7cm,connect spies,every spy on node/.append style={thick}}]
    \node {\includegraphics[height=0.25\textwidth]{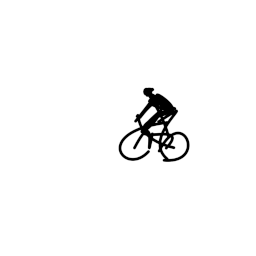}};
\end{tikzpicture}\\ 
\end{tabular} 
}
\caption{Failure cases}
\label{fig:failure}
\end{figure}
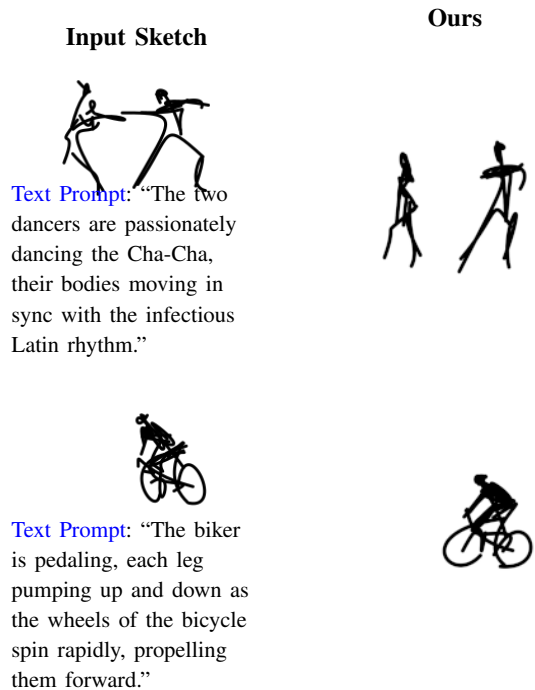


\section{Limitations}
Our method relies on a pre-trained text-to-video prior~\cite{zhu2023motionvideogan}, which may struggle with certain types of motion, leading to errors that propagate and manifest as noticeable artifacts in some cases of the generated animations. Improvements could be made by employing more advanced text-to-video priors capable of handling text-to-video alignment with higher accuracy. Additionally, our approach faces challenges in animating multi-object scenarios, particularly when functional relationships exist between objects. Designed primarily for single-object animations, the method experiences a decline in quality when dealing with such cases. For example, as shown in Figure~\ref{fig:failure}, the human and the bicycle are incorrectly separated, resulting in unnatural motion during the animation. Future work could address this limitation by implementing object-specific translations rather than relying on relative motion.


\section{Conclusion}
\label{sec:conclusion}
This work presents a method for generating animated sketches from a combination of sketch inputs and text prompts. To ensure temporal consistency in the animations, we introduce a Length-Area (LA) regularizer, and to preserve the original shape’s topology, we propose a shape-preserving ARAP loss. Our approach delivers superior performance both quantitatively and qualitatively, addressing challenges in animation generation. However, the method has certain limitations, including its inability to handle multi-object scenarios and its reliance on a pre-trained text-to-video prior. Future work will focus on addressing these limitations to further enhance the method’s capabilities.

\bibliography{reference.bib}
\bibliographystyle{ieeetr}

\end{document}